\documentclass[10pt,twocolumn,letterpaper]{article}
\usepackage{iccv}
\usepackage{times}
\usepackage{epsfig}
\usepackage{graphicx}
\usepackage{amsmath}
\usepackage{amssymb}
\usepackage[pagebackref=true,breaklinks=true,letterpaper=true,colorlinks,bookmarks=false]{hyperref}

\usepackage{multicol}
\usepackage{afterpage}
\usepackage{booktabs}
\usepackage{mathtools}
\usepackage{xfrac}
\usepackage[percent]{overpic}
\usepackage{xcolor}
\usepackage{multirow}
\usepackage{colortbl}
\usepackage{siunitx}
\usepackage{pifont}
\usepackage{subfig}
\usepackage[toc,page]{appendix}

\newcommand{\sect}[1]{Section~\ref{#1}}
\newcommand{\fig}[1]{Figure~\ref{#1}}
\newcommand{\figs}[1]{Figures~\ref{#1}}
\newcommand{\tab}[1]{Table~\ref{#1}}
\newcommand{\eq}[1]{Equation~\ref{#1}}

\newcommand{\ourpapertitle}{Zoom-to-Inpaint: Image Inpainting with High-Frequency Details}
\definecolor{mygray}{gray}{0.4}

\iccvfinalcopy

\ificcvfinal\fi

\begin{document}

\title{\ourpapertitle}

\author{Soo Ye Kim\textsuperscript{1,2}\footnotemark[2]
\qquad Kfir Aberman\textsuperscript{2} 
\qquad Nori Kanazawa\textsuperscript{2} 
\qquad Rahul Garg\textsuperscript{2} 
\qquad Neal Wadhwa\textsuperscript{2} \\ 
Huiwen Chang\textsuperscript{2} 
\qquad Nikhil Karnad\textsuperscript{2} 
\qquad Munchurl Kim\textsuperscript{1} 
\qquad Orly Liba\textsuperscript{2}
\\
[0.8em]
\begin{tabular}{*{2}{>{\centering}p{.35\textwidth}}}
\textsuperscript{1}KAIST & \textsuperscript{2}Google Research \tabularnewline
Daejeon, Republic of Korea & Mountain View CA, USA \tabularnewline
\end{tabular}
}

\maketitle
\footnotetext[2]{This work was done during an internship at Google Research.}
\begin{abstract}
  Although deep learning has enabled a huge leap forward in image inpainting, current methods are often unable to synthesize realistic high-frequency details. In this paper, we propose applying super-resolution to coarsely reconstructed outputs, refining them at high resolution, and then downscaling the output to the original resolution. By introducing high-resolution images to the refinement network, our framework is able to reconstruct finer details that are usually smoothed out due to spectral bias -- the tendency of neural networks to reconstruct low frequencies better than high frequencies. To assist training the refinement network on large upscaled holes, we propose a progressive learning technique in which the size of the missing regions increases as training progresses. Our zoom-in, refine and zoom-out strategy, combined with high-resolution supervision and progressive learning, constitutes a framework-agnostic approach for enhancing high-frequency details that can be applied to any CNN-based inpainting method. We provide qualitative and quantitative evaluations along with an ablation analysis to show the effectiveness of our approach. This seemingly simple, yet powerful approach, outperforms state-of-the-art inpainting methods. Our code is publicly available on the web.
\end{abstract}

\section{Introduction}

Image inpainting is a long-standing problem in computer
vision and has many graphics applications. The goal of the problem is to fill in missing regions in a masked image, such that the output is a natural completion of the captured scene with (i) plausible semantics, and (ii) realistic details and textures.
The latter can be achieved with traditional inpainting methods that copy patches of valid pixels, e.g., PatchMatch~\cite{barnes2009patchmatch}, thus preserving the textural statistics of the surrounding regions. Nevertheless, the inpainted results often lack semantic context and do not blend well with the rest of the image.
\begin{figure}[t]
\centering
\includegraphics[width=\columnwidth]{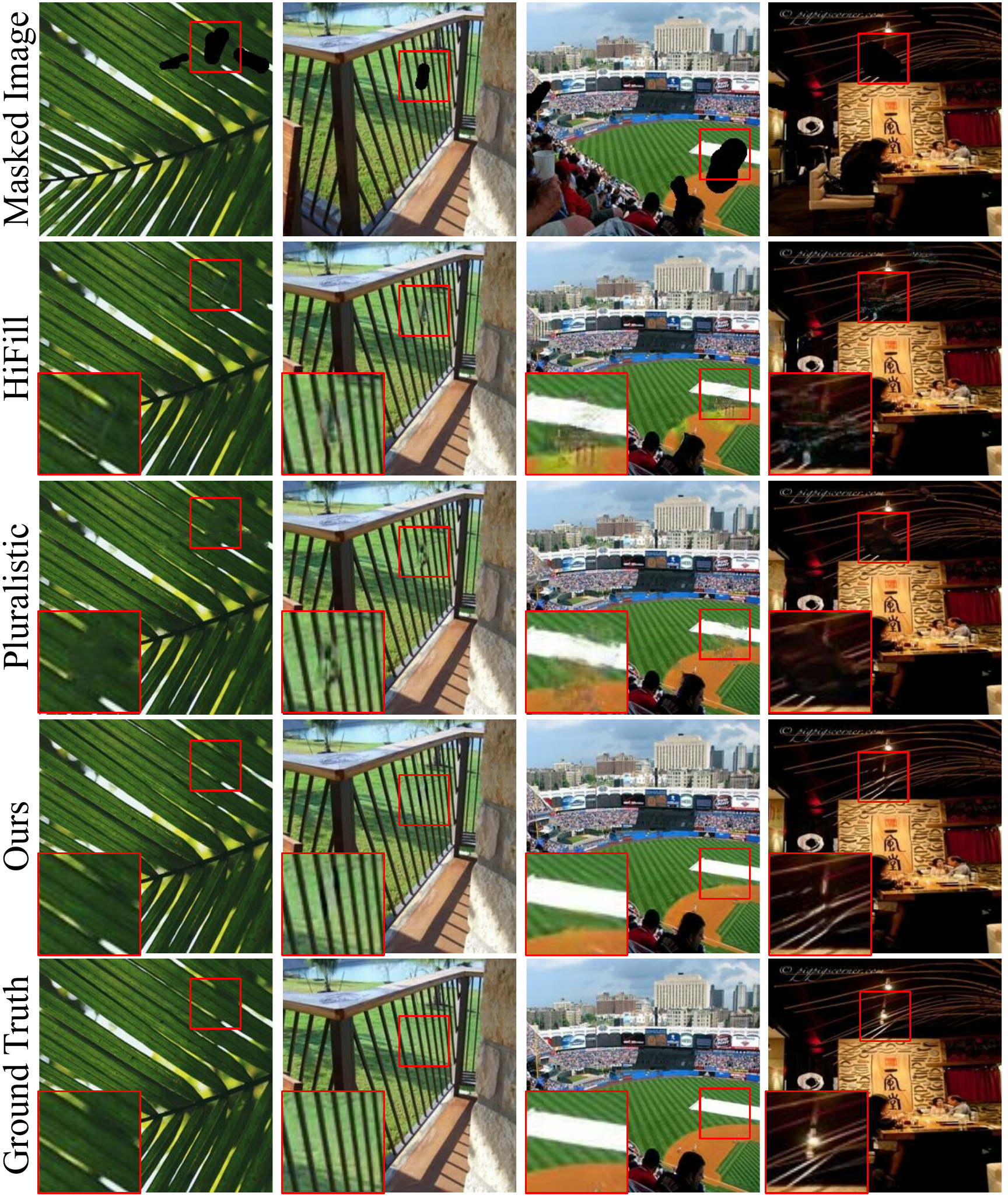}
\caption{Qualitative comparison to recent inpainting methods, HiFill \cite{yi2020contextual} and Pluralistic \cite{zheng2019pluralistic}. Our method correctly reconstructs high-frequency details, e.g, fine textures and narrow structures, and preserves the continuity and the orientation of edges.}
\vspace{-2mm}
\label{fig:teaser}
\end{figure}
With the advent of deep learning, inpainting neural networks are commonly trained in a self-supervised fashion, by generating random masks and applying them to the full image to produce masked images that are used as the network's input. These networks are able to produce semantically plausible results thanks to abundant training data. However, the results often do not have realistic details and textures, presumably due to the finding of a \textit{spectral bias} \cite{rahaman2019spectral} in neural networks. That is, high-frequency details are difficult to learn as neural networks are biased towards learning low-frequency components. This is especially problematic when training neural networks for image restoration tasks such as image inpainting, because high-frequency details must be generated for realistic results.

Recent neural network architectures for image inpainting consist of (i) a coarse network that first generates a coarsely filled-in result, and (ii) a refinement network that corrects and refines the coarse output for better quality \cite{yi2020contextual, yu2018generative, yu2019free}. In this paper, with the goal of generating more realistic high-frequency details, we propose to refine after \textit{zooming in}, therefore refining the image at a resolution higher than the target resolution. This allows the refinement network to correct local irregularities at a finer level and to learn from high-resolution (HR) labels, thus effectively reducing the spectral bias at the desired resolution and injecting more high-frequency details into the resulting image. We show that adding a simple bicubic upsampling component between the coarse and refinement networks improves inpainting results, and using a super-resolution (SR) network to upscale the intermediate result improves the results even further.

Furthermore, as an HR refinement network can be more difficult to train than a low-resolution (LR) refinement network due to a larger mask and more missing pixels in the HR input, we propose a novel \textit{progressive learning strategy} for inpainting, where the size of masks is increased as training progresses and the framework is trained on larger masks at a later training stage. Moreover, to further enhance the high-frequency details, we propose to use an additional gradient loss \cite{eigen2015predicting} that minimizes the gradients of the difference between the prediction and the ground truth. We believe that these three fundamental strategies can benefit any existing inpainting network.

\medskip
In summary, our contributions are as follows:
\begin{itemize}
    \item We propose a novel inpainting framework that includes an SR network to \textit{zoom in}, allowing refinement at HR and training with HR labels, to enhance the generation of high-frequency details in the final inpainted output. 
    \item We propose a progressive learning strategy for inpainting to aid convergence with larger masks.
    \item We use a gradient loss for inpainting to further improve textural details.
\end{itemize}

\begin{figure*}[t]
\centering
\includegraphics[width=\textwidth]{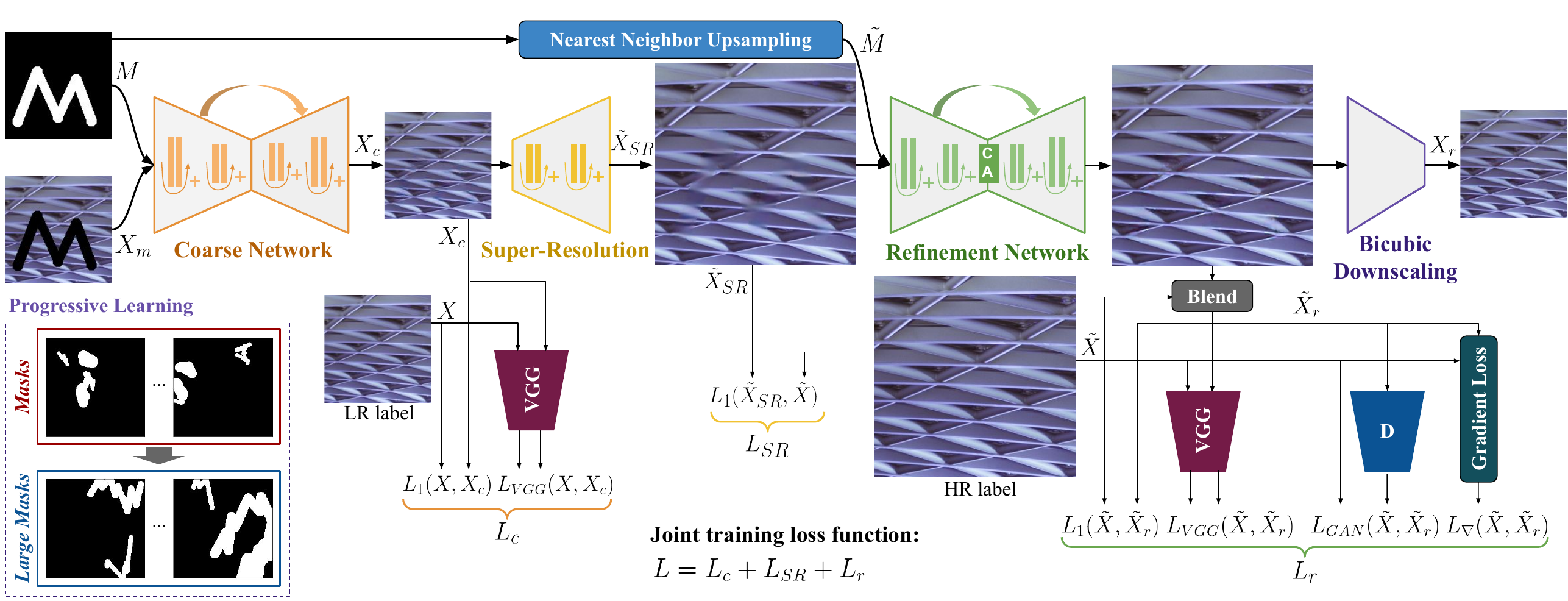}
\caption{Our proposed inpainting framework containing three main components: a coarse network, an SR network, and an HR refinement network. The framework progressively learns to inpaint larger missing regions, and the refinement network is trained with a gradient loss to enhance the generation of high-frequency details. Images shown in this figure are actual intermediate results produced by our framework, and more examples are given in the Appendix.
}
\vspace{-1mm}
\label{fig:framework}
\end{figure*}
\section{Related work}

\subsection{Image inpainting}
Traditional image inpainting methods can be largely classified into three types: (i) propagation-based approaches that gradually fill in the missing regions from known pixel values at hole boundaries \cite{bertalmio2000image, telea2004an}, (ii) Markov Random Field (MRF) approaches optimizing discrete MRFs \cite{komodakis2007image, pritch2009shift}, or (iii) patch-based approaches that search for plausible patches outside of the hole to be pasted into the missing region \cite{barnes2009patchmatch, barnes2010the} similar to texture synthesis algorithms \cite{efros2001image, efros1999texture}. These types of approaches exploit information already present in the input image.

Deep-learning-based inpainting methods leverage information external to any one specific image by learning global semantics from an abundant corpus of training data. An early convolutional neural network (CNN) based method for inpainting was the Context Encoder \cite{pathak2016context}, where the authors proposed using an L2 loss with a global generative adversarial network (GAN) loss for improved perceptual quality. GANs \cite{goodfellow2014generative} are especially suitable for image inpainting because they are able to synthesize realistic images \cite{brock2019large, karras2018progressive, radford2016unsupervised, zhang2017stackgan}. To consider local details as well as global semantics, Demir \textit{et al.} \cite{demir2018patchbased} proposed using a PatchGAN \cite{isola2017imagetoimage} along with the global GAN. Our inpainting framework also employs a PatchGAN discriminator for enhanced local details.

When generating each missing pixel, CNNs with stacked convolution layers are limited by the local receptive field of the convolution operation, whereas previous patch-based methods are able to copy from any part in the surrounding known regions. Thus, Yu \textit{et al.} \cite{yu2018generative} devised a contextual attention (CA) module that copies patches from the surrounding regions into the missing region, weighted by the computed similarity. We add the CA module in the bottleneck of our refinement network to get the best of both worlds -- ability of GANs in synthesizing novel structures and details, as well as the ability to copy patches anywhere from the image without restrictions in the receptive field, like in \cite{yu2018generative, yu2019free}. There also exist inpainting methods \cite{liu2019coherent, zeng2019learning} that use other types of attention modules.

Typically, input images for inpainting consist of some input pixels that are valid, or known, while others are invalid, or unknown/missing. Liu \textit{et al.} \cite{liu2018image} addressed this dichotomy using partial convolution, where only the valid pixels are taken into consideration during convolution by using a predefined mask. Yu \textit{et al.} \cite{yu2019free} proposed gated convolutions, where the masks are also learned, while Xie \textit{et al.} \cite{xie2019image} proposed using learnable bidirectional attention maps. We employ gated convolutions \cite{yu2019free} in our framework to handle the valid and invalid pixels.

Many recent CNN-based inpainting methods use a two-stage approach, where the first network generates a coarse output and the second network refines this output \cite{yi2020contextual, yu2018generative, yu2019free, zeng2020highresolution}. Some methods \cite{liao2018edge, nazeri2019edgeconnect, xiong2019foregroundaware} divide the stages as (i) edge completion and (ii) image completion, or in StructureFlow \cite{ren2019structureflow}, as (i) structure generation and (ii) texture generation.
Among the coarse-to-fine two-stage methods, some methods \cite{yi2020contextual, zeng2020highresolution} handle inpainting for HR images by first downscaling the HR input to a fixed resolution before the inpainting stages, and increasing the resolution after inpainting using residual aggregation \cite{yi2020contextual} or guided upsampling \cite{zeng2020highresolution}.
In our \textit{zoom-to-inpaint} model, we increase the resolution of the image \textit{between} the coarse inpainting and the refinement stages such that the coarse output is refined at HR, and then come back to the original or the desired resolution after refinement.

Unlike conventional CNN architectures for image and video restoration that tend to reduce the size of the image (as in multi-scale architectures \cite{kim2019fisr, nah2017deep}), or feature map (eg. U-Net \cite{ronneberger2015unet}), to increase the receptive field of the network, our model goes \textit{beyond} the input and target resolutions for better refinement of high-frequency details.
Upsampling can be achieved by simple bicubic interpolation, and we show that using an SR network, trained end-to-end with the coarse and refinement inpainting networks, produces even better results.  
The modular construction of this seemingly simple idea is in fact non-trivial, and we explain the framework in detail in \sect{sect:method}. We further elaborate on how HR refinement aids the generation of high-frequency details in \sect{HR_refinement} and present a frequency-based analysis in \sect{sec:ablation}.

\subsection{Progressive learning}
Learning to solve a difficult task from scratch can be challenging for neural networks. Hence, fine-tuning \cite{hinton504} or transfer learning is a commonly applied technique, where prior knowledge is transferred from pretrained networks to the subsequent training stages. Progressive neural networks \cite{rusu2016progressive} expanded on this idea and added lateral connections to previously learned features. Karras \textit{et al.} \cite{karras2018progressive} proposed to progressively train a GAN to synthesize LR images first, then to generate HR images by incrementally adding on layers to stabilize and speed up the training process. For image inpainting, filling in missing regions can be increasingly difficult as they become larger. Hence, some methods \cite{guo2019progressive, li2020recurrent} proposed a recurrent scheme that iteratively fills in holes from the boundary for each image as in traditional propagation-based schemes, which must also be applied during inference. In contrast, we propose to increase the size of the masks as training progresses so that our inpainting network \textit{learns} progressively.

\subsection{Gradient loss}
Utilizing image gradients as a prior \cite{sun2008image} or in the loss function \cite{eigen2015predicting, li2018megadepth, ma2020structure} has been widely explored for image SR \cite{ma2020structure, sun2008image} and depth estimation \cite{eigen2015predicting, li2018megadepth} to increase sharpness in the reconstructed images. In image inpainting, Telea \textit{et al.} \cite{telea2004an} proposed a fast marching method that use the gradients of neighboring pixels to estimate the missing values in the inpainted region. Liu \textit{et al.} \cite{liu2018imageb} proposed a loss function to enforce the continuity of gradients in the reconstructed region and its neighboring regions. Inspired by Eigen \textit{et al.}'s method \cite{eigen2015predicting} for depth estimation, we propose to minimize the image gradients of the difference between the prediction and the ground truth to further enhance high-frequency details in the inpainted result.

\section{Proposed method}
\label{sect:method}

We propose a novel inpainting framework that is able to reconstruct high-frequency details in the final output by (i) upsampling the result of a coarse inpainting network
using an SR network and refining at HR, and (ii) employing a gradient loss. For better convergence, the framework is trained \textit{progressively}, by increasing the size of masks.

\subsection{Framework overview} \label{sect:overview}
Our inpainting framework consists of three trainable networks connected sequentially: a coarse inpainting network, an SR network, and an HR refinement network. The SR network and the HR refinement network are trained with HR (original) labels, $\tilde{X}$, whereas the coarse network is trained with the bicubic-downscaled versions, $X$. Each network is first pretrained separately, and then all networks are trained jointly. Please refer to the Appendix for the detailed architectures of each network.

\subsubsection{Coarse network}
The coarse network, $f_c$, aims to characterize the LR variations in the image across its entire field-of-view, coarsely filling in the missing regions in the input masked image, $X_m$, given by,

\begin{equation} \label{eq:masked_image}
    X_m = (1-M)\odot X,
\end{equation}
where $M\in\mathbb{R}^{H\times W\times 3}$ is a binary mask where invalid pixels, i.e., pixels to be inpainted, are 1 and valid pixels are 0, with repeated values across the channel dimension, $X\in\mathbb{R}^{H\times W\times 3}$ is the full image, and $\odot$ is element-wise multiplication. 
We employ an encoder-decoder-based CNN architecture with gated convolutions similar to \cite{yu2019free}, additionally with residual blocks and batch normalization.
The network output is masked with the input, yielding $X_c$ as, 
\begin{equation} \label{eq:coarse_output}
    X_c = M\odot f_c(X_m, M, \Theta_c) + X_m,
\end{equation}
so that the network does not attempt to reconstruct already valid regions. We train it by minimizing a loss function $L_c$, consisting of an L1 loss to enforce pixel-wise similarity, and a VGG loss to enforce similarity in the feature domain, given as,
\begin{equation}
    L_c = {\lVert X_c - X\rVert}_1 + \lambda_c^{\phi}\cdot {\lVert\phi_{1, 4}(X_c) - \phi_{1, 4}(X)\rVert}_1,\label{eq:Lc}
\end{equation} 
where $\phi_{i, j}$ is the $i$-th convolution layer at the $j$-th block in VGG19 \cite{simonyan15}, and $\lambda_c^{\phi}$ is a constant. 

\subsubsection{Super-resolution network}
\label{sec:super_res}
We use an SR network that \textit{zooms in} on the coarse output $X_c$ by scale factor $s > 1$, yielding $\tilde{X}_{SR}\in\mathbb{R}^{sH\times sW\times 3}$. Our SR network architecture is designed as a cascade of four residual blocks with a pixel shuffle layer \cite{shi2016realtime} at the end.
Contrary to the coarse network output, we do not mask $\tilde{X}_{SR}$ since the refinement network in the following stage can propagate the HR patches from valid regions to the inpainted region using contextual attention (CA) \cite{yu2018generative}.
Therefore, we train the SR network by directly minimizing $L_{SR}={\lVert \tilde{X}_{SR} - \tilde{X}\rVert}_1$, where $\tilde{X}\in\mathbb{R}^{sH\times sW\times 3}$ is the full HR image.

\subsubsection{High-resolution refinement network}
\label{HR_refinement}
Unlike common refinement schemes of previous inpainting frameworks, our proposed refinement is achieved by \textit{zooming in}, refining, then \textit{zooming out} back to the input resolution, in order to benefit from the supervision of HR labels during refinement and aid the learning of high-frequency components.
Specifically, given $\tilde{X}_{SR}$ and $\tilde{M}$ as input, where $\tilde{M}\in\mathbb{R}^{sH\times sW\times 3}$ is $M$ upscaled by nearest neighbor upsampling, the HR refinement network, $f_r$, generates the refined image $f_r(\tilde{X}_{SR}, \tilde{M}, \Theta_r)\in\mathbb{R}^{sH\times sW\times 3}$. Then, $\tilde{X}_r$, which is used for optimizing the training losses in the refinement network, is obtained by blending the network output with the label, $\tilde{X}$:
\begin{equation} \label{eq:refinement_output}
    \tilde{X}_r = \tilde{M}\odot f_r(\tilde{X}_{SR}, \tilde{M}, \Theta_r) + (1-\tilde{M})\odot \tilde{X}.
\end{equation}
By masking with the original label and not the input, $\tilde{X}_{SR}$, no loss occurs outside the inpainted regions. As the loss is zero in the valid regions, the network does not spend its capacity on reconstructing these regions, which will be replaced by the input image, $X_m$, after downscaling, in the final output (\eq{eq:downscaling}). The architecture of the refinement network is similar to the coarse network and is encoder-decoder-based, except that we add a CA module \cite{yu2018generative} to its bottleneck.

For training, we use a gradient loss, $L_\nabla$, between $\tilde{X}_r$ and $\tilde{X}$ to further encourage the generation of high-frequency details. Inspired by \cite{eigen2015predicting}, $L_\nabla$ is given as,
\begin{equation}
    L_\nabla = \frac{1}{2}({\lVert(\tilde{X}_r - \tilde{X})_{\nabla_x}\rVert}^2_2+{\lVert(\tilde{X}_r - \tilde{X})_{\nabla_y}\rVert}^2_2),
\end{equation}
where $\nabla_x$ and $\nabla_y$ are horizontal and vertical image gradients, respectively, obtained by 1-tap filters, $[-1, 1]$ and ${[-1, 1]}^\text{T}$. Then, $f_r$ is trained with $L_r$ that consists of an L1 loss, VGG loss, hinge GAN loss -- $L_h$, and $L_\nabla$, given by,
\begin{align} \label{eq:loss_refine}
    L_r = {\lVert \tilde{X}_r - \tilde{X}\rVert}_1 &+ \lambda_r^{\phi}\cdot {\lVert\phi_{1, 4}(\tilde{X}_r) - \phi_{1, 4}(\tilde{X})\rVert}_1 \notag\\
    &+ \lambda_r^{GAN}\cdot L_h(\tilde{X}_r) + \lambda_r^{\nabla}\cdot L_\nabla.
\end{align}
A PatchGAN \cite{isola2017imagetoimage} approach is adopted for good perceptual results, with spectral normalization \cite{miyato2018spectral} similar to \cite{yu2019free} for stable training of GANs. 
By training the refinement network with HR labels, we drive the CNN to explicitly learn high-frequency details, additionally to the low-frequencies that are inherently preferred according to the empirical evidence of a spectral bias in neural networks \cite{rahaman2019spectral}.

After pretraining each of the three components separately, the entire framework is trained end-to-end with a total loss $L=L_c+L_{SR}+L_r$.

\medskip\noindent
\textbf{Downscaling.}\quad
As a last step, the refined HR output, $f_r(\tilde{X}_{SR}, \tilde{M}, \Theta_r)$, is downscaled back (\textit{zoomed out}) by scale factor $s$ to the original resolution by bicubic downsampling, and blended with the input, $X_m$.
The final output $X_r\in\mathbb{R}^{H\times W\times 3}$ is then given as,
\begin{equation}
    X_r = M\odot f_r(\tilde{X}_{SR}, \tilde{M}, \Theta_r)\downarrow_s + (1-M)\odot X_m. \label{eq:downscaling}
\end{equation}
Note that the refinement network would not be able to learn from the HR labels if losses are only imposed on $X_r$.

\begin{figure}[t]
\centering
\includegraphics[width=\columnwidth]{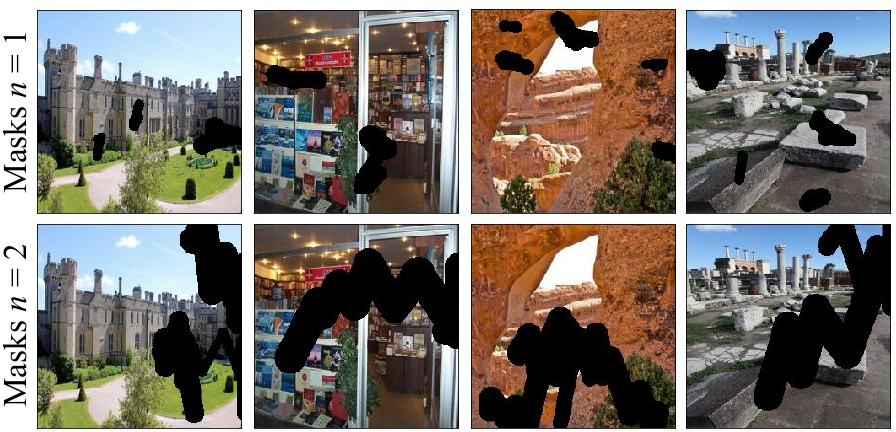}
\caption{Example masks used during progressive learning.}
\label{fig:prog_masks}
\end{figure}

\subsection{Progressive learning for image inpainting} \label{sect:progressive_inpainting}
In image inpainting, the training time until convergence tends to increase proportionally as invalid regions in masks become larger and increasingly more difficult to fill in \cite{liu2018image}.
This is problematic if we want to refine at HR with the image enlarged by scale factor $s$, since the number of missing pixels would increase by $s^2$. Thus, we propose a progressive learning strategy for inpainting, where we train the network in $N$ steps by increasing the size of masks at each $n$-th step, where $n=1, 2, ..., N$. We set $N=2$ in our experiments, where masks at $n=2$ are the same as random masks used in \cite{yu2018generative} and masks at $n=1$ are generated by modifying the random generation parameters of masks in \cite{yu2018generative} to produce smaller and more confined masks. Example masks are shown in \fig{fig:prog_masks} and a detailed configuration of the parameters is provided in the Appendix.

Empirically, if our framework is trained directly on masks at $n=2$ without progressive learning, it takes 2M iterations to converge. With progressive learning, it takes 80K iterations to converge on masks at $n=1$ then only 1.2M iterations on masks at $n=2$. A mere addition of 80K iterations on $n=1$ drops the total number of iterations by $\sim40\%$. In the following sections, \textit{masks} denote masks at $n=1$ and \textit{large masks} denote masks at $n=2$.

\section{Experiment results and evaluation} 
\label{sect:experiments}

\subsection{Implementation details}\label{sect:implementation}
\noindent
\textbf{Training configuration.}\quad
For training our \textit{zoom-to-inpaint} model, we first pretrain the coarse and refinement networks on Places2 \cite{zhou2017places} at $256\times 256$ resolution for 180K iterations using \textit{masks} as defined in \sect{sect:progressive_inpainting}. For both networks, we only use the L1 loss and the VGG loss, with $\lambda^{\phi}=0.01$. For the SR network, we use an upsampling scale factor of $s=2$ and pretrain it on DIV2K \cite{agustsson2017ntire} for 400K iterations with randomly cropped $64\times 64$ patches. Then, we jointly train the entire framework on DIV2K using the proposed progressive learning strategy, i.e., 80K iterations with \textit{masks}, and then another 1.2M iterations with \textit{large masks}. We randomly crop $512\times 512$ patches from DIV2K and use them as HR labels ($\tilde{X}$), and bicubic-downsample them to $256\times 256$ to generate LR labels ($X$). We use the following loss coefficients: $\lambda_c^{\phi}=0.01$, $\lambda_r^{\phi}=10^{-5}$, $\lambda_r^{GAN}=0.5$ and $\lambda_r^{\nabla}=1$. 
Our implementation is in Tensorflow \cite{tensorflow2015-whitepaper} and trained on  8 NVIDIA V100 GPUs using Adam optimizer \cite{kingma2017adam} with a mini-batch size of 16 and a learning rate of $10^{-5}$.
Please refer to the Appendix for details of our model architecture and a complexity analysis.

\medskip\noindent
\textbf{Test dataset.}\quad
For the test dataset, we apply both masks on 200 images from the validation and test sets of Places2 (100 images each), and on 100 images in the validation set of DIV2K. For DIV2K, $256\times256$ patches were center-cropped from the full images and used as $X$. 

\begin{table*}
    \centering
    \scalebox{0.9}{
    \begin{tabular}{c|c|c|c|c||c|c|c|c}
         \multirow{2}{*}{Method} & \multicolumn{4}{c||}{Places2 ($256\times256$)} & \multicolumn{4}{c}{DIV2K ($256\times256$, center-cropped)} \\
         \cline{2-9}
          & PSNR $\uparrow$ & SSIM $\uparrow$ & MS-SSIM $\uparrow$ & L1 Error $\downarrow$ & PSNR $\uparrow$ & SSIM $\uparrow$ & MS-SSIM $\uparrow$ & L1 Error $\downarrow$\\
         \hline
         \rowcolor[gray]{.95}\multicolumn{9}{c}{\textit{Masks}} \\
         \hline
         HiFill \cite{yi2020contextual} & 31.12 & 0.9586 & 0.9742 & 0.00744 & 30.91 & 0.9633 & 0.9777 & 0.00700 \\
         \hline
         Pluralistic \cite{zheng2019pluralistic} & 33.23 & 0.9670 & 0.9807 & 0.00558 & 32.73 & 0.9703 & 0.9820 & 0.00543 \\
         \hline
         DeepFill-v2 \cite{yu2019free} & 34.03 & 0.9719 & 0.9834 & 0.00485 & 33.11 & 0.9741 & 0.9852 & 0.00467 \\
         \hline
         EdgeConnect \cite{nazeri2019edgeconnect} & 33.98 & 0.9718 & 0.9841 & 0.00388 & 33.11 & 0.9734 & 0.9845 & 0.00388 \\
         \hline\hline
         Ours & \textbf{34.78} & \textbf{0.9755} & \textbf{0.9863} & \textbf{0.00357} & \textbf{34.08} & \textbf{0.9787} & \textbf{0.9886} & \textbf{0.00329} \\
         \multicolumn{9}{c}{\,} \\
         \hline
         \rowcolor[gray]{.95}\multicolumn{9}{c}{\textit{Large Masks}} \\
         \hline
         HiFill \cite{yi2020contextual} & 24.94 & 0.8891 & 0.9134 & 0.02034 & 24.23 & 0.8739 & 0.8993 & 0.02302 \\
         \hline
         Pluralistic \cite{zheng2019pluralistic} & 26.17 & 0.9022 & 0.9191 & 0.01784 & 25.62 & 0.8890 & 0.9071 & 0.01949 \\
         \hline
         DeepFill-v2 \cite{yu2019free} & 26.77 & 0.9158 & 0.9326 & 0.01536 & 26.07 & 0.9018 & 0.9229 & 0.01735 \\
         \hline
         EdgeConnect \cite{nazeri2019edgeconnect} & 27.61 & 0.9166 & 0.9382 & 0.01328 & 26.87 & 0.9036 & 0.9291 & 0.01494 \\
         \hline\hline
         Ours & \textbf{27.71} & \textbf{0.9202} & \textbf{0.9415} & \textbf{0.01314} & \textbf{27.07} & \textbf{0.9094} & \textbf{0.9346} & \textbf{0.01462} \\
    \end{tabular}}
    \caption{Quantitative evaluation. Values in \textbf{bold} denote the best performance.}
    \label{tab:quantitative}
\end{table*}

\begin{table}
    \centering
    \scalebox{0.88}{
    \begin{tabular}{c|c|c|c|c}
        Compared method & \cite{yi2020contextual} & \cite{zheng2019pluralistic} & \cite{yu2019free} & \cite{nazeri2019edgeconnect} \\
        \hline\hline
        Preference of & \multirow{2}{*}{75.49\%} & \multirow{2}{*}{89.13\%} & \multirow{2}{*}{69.23\%} & \multirow{2}{*}{64.21\%} \\
        ours over compared &&&&\\
    \end{tabular}}
    \caption{User study results on \textit{masks}, indicating the percentage (\%) of users who selected our method over the compared method. Results on \textit{large masks} are provided in the Appendix.}
    \label{tab:user_study}
    \vspace{-1mm}
\end{table}

\medskip\noindent
\textbf{Inpainting methods.}\quad
We compare our method with recent state-of-the-art inpainting approaches: DeepFillv2 \cite{yu2019free}, EdgeConnect \cite{nazeri2019edgeconnect}, Pluralistic \cite{zheng2019pluralistic}, and HiFill \cite{yi2020contextual}. 
Similar to our method, \cite{nazeri2019edgeconnect, yu2019free,zheng2019pluralistic} were trained on $256\times256$ images, and therefore, our test set can be used as is.
However, since HiFill \cite{yi2020contextual} was originally trained on $512\times512$ images, we bicubic-upscale the test images to $512\times512$ for input and then downscale the output back to $256\times256$ before computing the metrics. We used the publicly released weights trained on Places2 for all methods. 

\subsection{Quantitative evaluation}
For quantitative evaluation, we report the results of four metrics -- PSNR, SSIM \cite{wang2004image}, MS-SSIM (multi-scale SSIM) \cite{wang2003multiscale} and L1 error -- that are frequently used in inpainting literature, on both mask sizes in \tab{tab:quantitative}. Two additional perceptual metrics -- FID and LPIPS -- that are less frequently used, are reported in the Appendix.
As shown in \tab{tab:quantitative}, our \textit{zoom-to-inpaint} model outperforms all compared methods on both mask sizes on all metrics, with at most 0.97 dB PSNR and 0.0055 MS-SSIM gain over the next best method. It shows that our framework is able to generate results that are more consistent with the ground truth compared to the recent state-of-the-art inpainting methods. 

In \tab{tab:quantitative}, the improvement over the next best method on PSNR and L1 error, which are pixel-wise error metrics, is larger on \textit{masks} compared to \textit{large masks}, showing that our framework adds on better pixel-level details that are closer to the ground truth when the missing region is relatively smaller. The gain on MS-SSIM, which measures the structural similarity at multiple scales, is greater for \textit{large masks}, showing that our model is able to recover better global structures for large missing regions.

\subsection{Qualitative evaluation}

\begin{figure*}
    \centering
    \includegraphics[width=\textwidth]{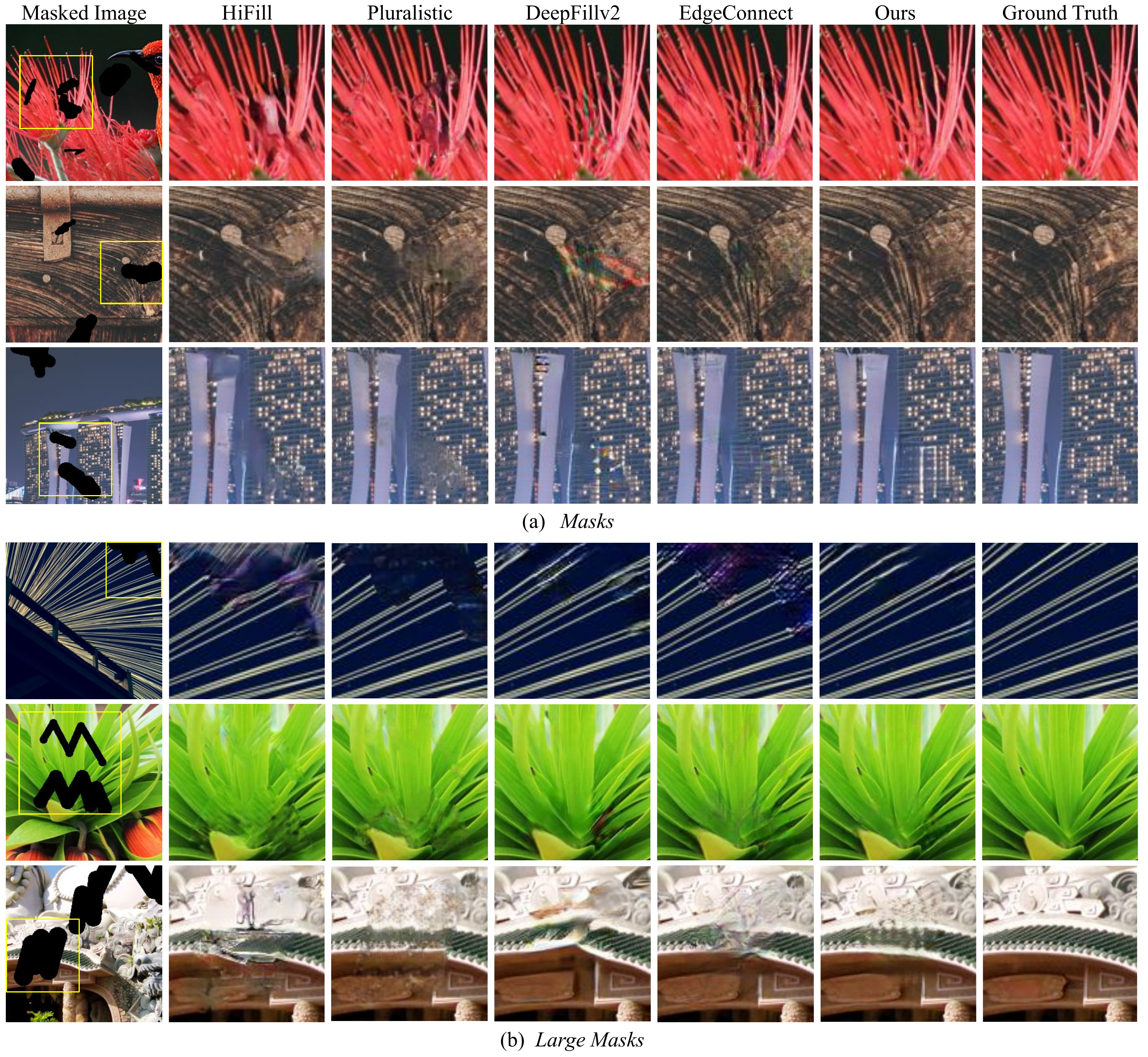}
    \caption{Qualitative comparison to other methods on (a) \textit{Masks} and (b) \textit{Large Masks}. When images with high-frequency regions are fed into inpainting networks, HiFill \cite{yi2020contextual} and Pluralistic \cite{zheng2019pluralistic} tend to generate blurry texture as seen in the examples in the 1st, 2nd and 3rd rows, and DeepFillv2 \cite{yu2019free} and EdgeConnect \cite{nazeri2019edgeconnect} generate color artifacts as shown in the 2nd, 3rd and 5th rows. Our method is able to accurately reconstruct high-frequency details, as well as global structures.
    }
    \label{fig:qualitative_1}
\end{figure*}

\noindent
\textbf{Visual results.}\quad
We show qualitative comparisons of visual results in \figs{fig:teaser} and~\ref{fig:qualitative_1}. 
In \fig{fig:teaser}, our method accurately reconstructs edges and fine lines in the correct orientation while
other methods find it difficult to preserve the continuity of the fine lines or fail to produce any edges at all in the missing region.
\fig{fig:qualitative_1} shows the qualitative results on both mask sizes. Similar to \fig{fig:teaser}, our method accurately reconstructs high-frequency details such as edges and fine texture, as well as global structures for \textit{large masks}.
Please refer to the Appendix for additional results, including full images of the crops shown in \fig{fig:qualitative_1}.

\begin{table}
    \centering
    \scalebox{0.85}{
    \begin{tabular}{l|c|c|c|c}
        Ablations & PSNR $\uparrow$ & SSIM $\uparrow$ & MS-SSIM $\uparrow$ & L1 Error $\downarrow$ \\
        \hline
        \rowcolor[gray]{.95}\multicolumn{5}{c}{\textit{Masks}} \\
        \hline
        No zoom & 32.12 & 0.9714 & 0.9812 & 0.00441\\
        \hline
        Bicubic zoom & 32.80 & 0.9753 & 0.9832 & 0.00391\\
        \hline
        SR zoom & 33.40 & 0.9770 & 0.9853 & 0.00363 \\
        \hline
        SR zoom+$L_\nabla$ & \textbf{34.08} & \textbf{0.9787} & \textbf{0.9886} & \textbf{0.00329} \\
        \multicolumn{5}{c}{\,} \\
        \hline
        \rowcolor[gray]{.95}\multicolumn{5}{c}{\textit{Large Masks}} \\
        \hline
        No zoom & 25.89 & 0.8977 & 0.9180 & 0.01747\\
        \hline
        Bicubic zoom & 26.09 & 0.9016 & 0.9219 & 0.01657\\
        \hline
        SR zoom & 26.95 & 0.9080 & 0.9307 & 0.01497 \\
        \hline
        SR zoom+$L_\nabla$ & \textbf{27.07} & \textbf{0.9094} & \textbf{0.9346} & \textbf{0.01462} \\
    \end{tabular}}
    \caption{Quantitative comparison of our model (SR zoom+$L_\nabla$) with its ablations.}
    \vspace{-1mm}
    \label{tab:ablation}
\end{table}

\medskip\noindent
\textbf{User study.}\quad
We conduct a user study to evaluate the preferences of users on the results produced by our approach compared to the other methods. We asked 13 users to evaluate 300 pairs of inpainted images in a random order, where one image is generated by our method, and the other image is generated by a method among \cite{nazeri2019edgeconnect, yi2020contextual, yu2019free, zheng2019pluralistic}.
Users are asked to select their preferred result based on the question: ``Which of these images looks better?". The percentage of users that prefer our method over the others for \textit{masks} is summarized in \tab{tab:user_study}, where users more frequently prefer our method over all other methods, with at least 64.21\% preference rate. More details of the user study including a screenshot, the raw number of counts, and results on \textit{large masks} are provided in the Appendix. We observed that \textit{masks} are more suitable for comparing the ability to generate pleasing and comparable results rather than \textit{large masks}, due to objectionable artifacts being generated by all methods for the latter, as shown in the Appendix.

\subsection{Ablation study} \label{sec:ablation}

\begin{figure}
    \centering
    \includegraphics[width=\columnwidth]{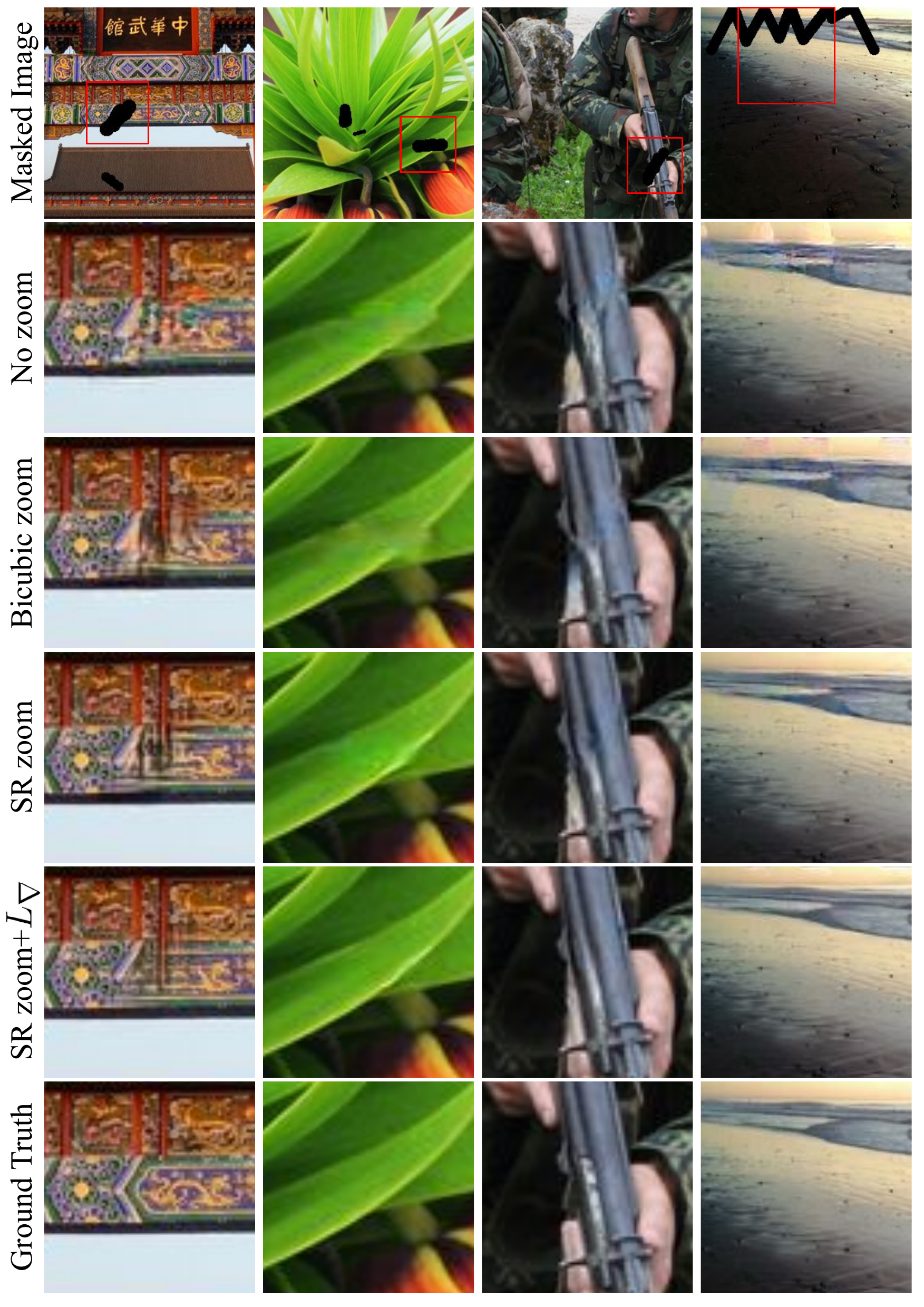}
    \caption{Visual comparison of results produced by our model and its ablations. HR refinement and the gradient loss improves the generation of high-frequency details.}
    \label{fig:ablation}
\end{figure}

\noindent
\textbf{Ablation study on framework components.}\quad
In order to analyze the contributions of the individual components, we compare against three ablations of our inpainting framework: (i) No zoom, (ii) Bicubic zoom, and (iii) SR zoom. For (i), we replace the SR component (described in Section~\ref{sec:super_res}) with an identity transform that simply copies the coarse output without an upsampling component, so that refinement is applied at the original resolution like in other conventional two-stage inpainting frameworks. For (ii), we replace the SR component with bicubic upsampling, and for (iii), we add back the SR zoom. (i), (ii) and (iii) are trained without the gradient loss $L_\nabla$. Lastly, SR zoom+$L_\nabla$ corresponds to our full framework with $L_\nabla$. The results are shown in \tab{tab:ablation} and \fig{fig:ablation}.

As shown in \tab{tab:ablation}, zooming in with bicubic upsampling improves all metrics compared to refining at the original resolution (\emph{No zoom}), showing the benefit of refining at a higher resolution and training the refinement network with HR supervision.
This indicates that as long as the refinement network is trained on HR labels so that local irregularities generated by the coarse network can be corrected with the magnification, the \textit{zoom} can even be achieved by bicubic upsampling.
Adding the SR network further improves the accuracy by a large margin, with $>$1 dB gain in PSNR compared to \textit{No zoom}.
Compared to Bicubic zoom, SR zoom is able to generate sharper results for the surrounding regions, that can then be propagated by the CA module into the inpainted region.
Adding the gradient loss further improves the quantitative metrics for both mask sizes.
Its benefit is especially prominent for \textit{masks}, where the network is more likely to reconstruct the image more accurately, and thus, fine details contribute more to the evaluation metrics.
\fig{fig:ablation} shows that each component improves the reconstruction quality, analogously to the quantitative results. Please refer to the Appendix for additional results.

\medskip\noindent
\textbf{Frequency-domain analysis.}\quad
We provide insights into the benefits of zooming in and refining at HR even though the final output is of lower resolution. Using a frequency-domain analysis, we demonstrate that our strategy introduces desirable frequencies into the inpainted result that survive downsampling. Specifically, instead of directly computing the metrics corresponding to a ground truth image and a prediction, we first construct a 2-level Laplacian pyramid \cite{burt1983laplacian} for each of them using a traditional 5-tap Gaussian kernel, and report \textit{per-level} metric values. This allows us to measure the accuracy in different frequency bands.

\begin{figure}
    \centering
    \includegraphics[width=\columnwidth]{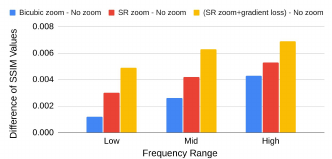}
    \caption{Frequency domain comparison of the ablation models using 2-level Laplacian pyramids \cite{burt1983laplacian}, evaluated on \textit{masks}. Each component of our framework adds to improving the reconstruction of the inpainted regions, with high frequencies benefitting more than lower frequencies.}

    \label{fig:freq_analysis}
\end{figure}

In \fig{fig:freq_analysis}, we show the per-level improvement of \emph{Bicubic zoom}, \emph{SR zoom} and \emph{SR zoom+$L_{\nabla}$} over the baseline \emph{No zoom}. We use the SSIM metric that is known to be sensitive to local structural changes, and use \textit{masks} to avoid the effect of artifacts generated with very large masks. While we see improvements in all frequency bands, we observe that the improvements are skewed towards higher frequency bands for all models. This indicates that all components of our framework, i.e., refining at HR with HR labels, SR zoom and gradient loss, improve the overall reconstruction, more so at higher frequencies. Please refer to the Appendix for additional results.

\section{Conclusion}
We propose a novel inpainting framework with HR refinement, by inserting an SR network between coarse and refinement networks. By training the refinement network with HR labels, our model is able to learn from high-frequency components present in the HR labels, reducing the spectral bias \cite{rahaman2019spectral} at the desired resolution.
Furthermore, we propose a progressive learning strategy for inpainting that increases the area of the missing regions as training progresses, and a gradient loss for inpainting to generate even more accurate texture and details. The HR refinement, progressive learning and gradient loss can each or together be applied to any inpainting framework. These simple but non-trivial modular constructions greatly improve the final inpainted result quantitatively and qualitatively. Our code is publicly available on the web.

\medskip\noindent
\textbf{Limitations.}\quad
In cases of challenging inputs with very large holes where all methods tend to generate severe artifacts, we find that our method produces artifacts containing a high-frequency repetitive pattern that is displeasing and sometimes more objectionable than artifacts produced by other methods. We analyze these artifacts with the user study on \textit{large masks} in the Appendix.

\paragraph{Acknowledgement}
We thank Jonathan T. Barron for insightful discussions and Yael Pritch and David Salesin for useful comments.

{\small
\bibliographystyle{ieee_fullname}

}

\begin{appendices}
\section{Detailed network architectures}
In this section, we give details of the network architectures of the different components in our zoom-to-inpaint framework.

\subsection{Coarse network}
\begin{figure*}
\centering
\includegraphics[width=\textwidth]{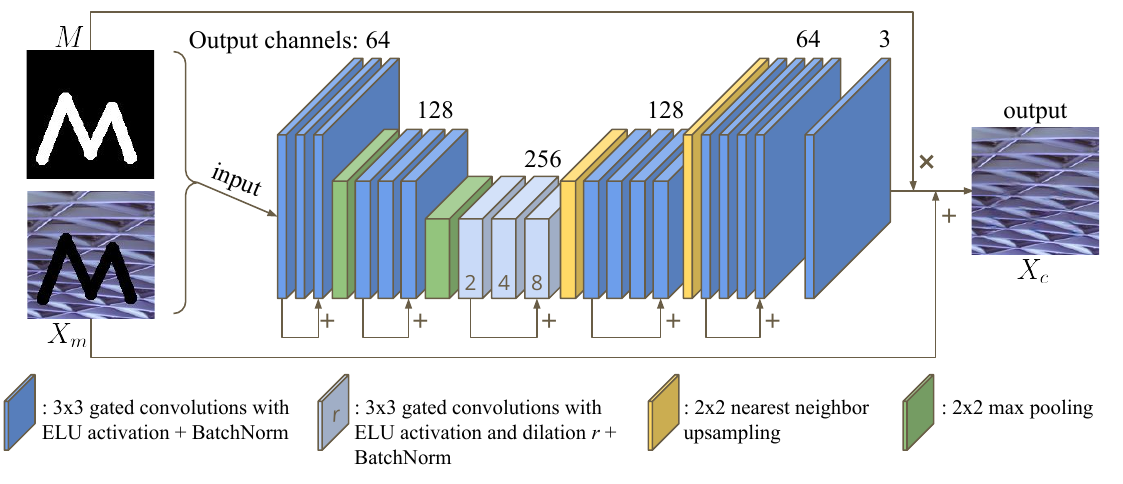}
\caption{Architecture of the coarse inpainting network in our zoom-to-inpaint framework.}
\label{fig:coarse_net}
\end{figure*}

The coarse network in our framework has an encoder-decoder-based architecture with gated convolutions \cite{yu2019free} and ELU activation \cite{clevert2016fast}. Different from the coarse network of \cite{yu2019free}, we add residual connections at each encoder and decoder level, and insert batch normalization (BatchNorm) \cite{ioffe2015batch} to every layer. We find that the residual connections help to propagate the information at each level and help convergence in training. For downscaling the feature resolution, we use max pooling with a $2\times 2$ window, and for upscaling we use nearest neighbor upsampling by a factor of 2 with an additional convolution layer to avoid checkerboard artifacts \cite{odena2016deconvolution}. At the bottleneck, the gated convolutions are applied with dilation rate set to 2, 4 and 8 like in \cite{yu2019free}, for an enlarged receptive field. All gated convolution filters are of size $3\times 3$, and the number of output channels is denoted at the top of each level in \fig{fig:coarse_net}. The network output is computed by blending with the input as mentioned in \eq{eq:coarse_output}.

\subsection{Super-resolution network}
\begin{figure}
\centering
\includegraphics[width=\columnwidth]{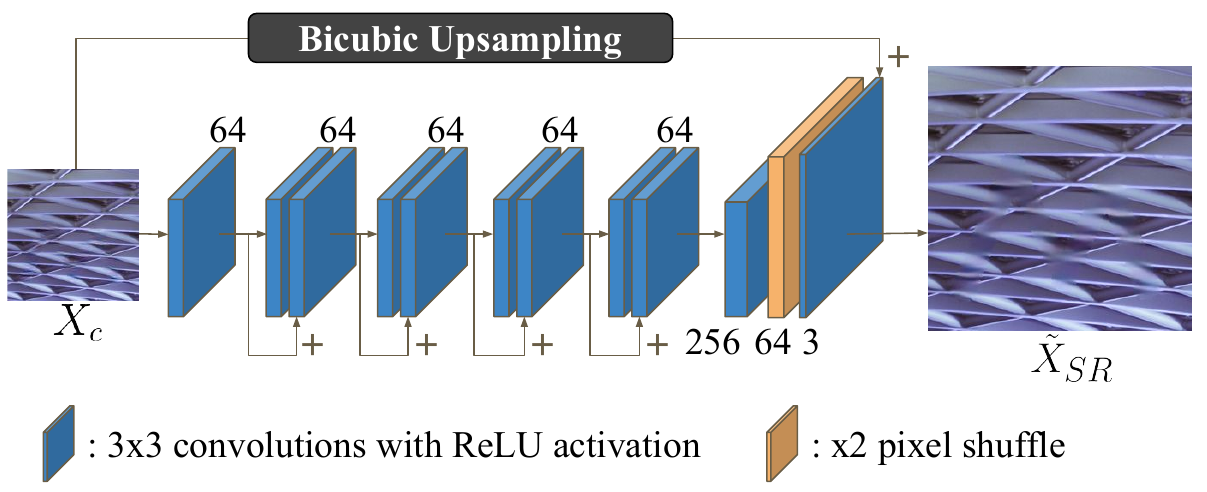}
\caption{Architecture of the SR network in our zoom-to-inpaint framework.}
\label{fig:sr_net}
\end{figure}

The super-resolution (SR) network in our framework is shown in \fig{fig:sr_net}. It has four cascaded residual blocks, each composed of two convolution layers with ReLU activation \cite{glorot2011deep}. We employ pixel shuffle \cite{shi2016realtime} at the end so that most computations are performed at low resolution (LR), to increase the size of the effective receptive field of the network, as well as to reduce computational complexity. All output channels are 64 except for the layer before pixel shuffle, which is 256, and the last convolution layer, which is 3, for reconstructing RGB channels for the output. We add a global residual connection with bicubic upsampling as in \cite{kim2016accurate} so that the network can focus on recovering the missing high-frequency components rather than on low-frequency components that are already present in the input. All convolution filters are of size $3\times 3$.

\begin{figure}
\centering
\includegraphics[width=\columnwidth]{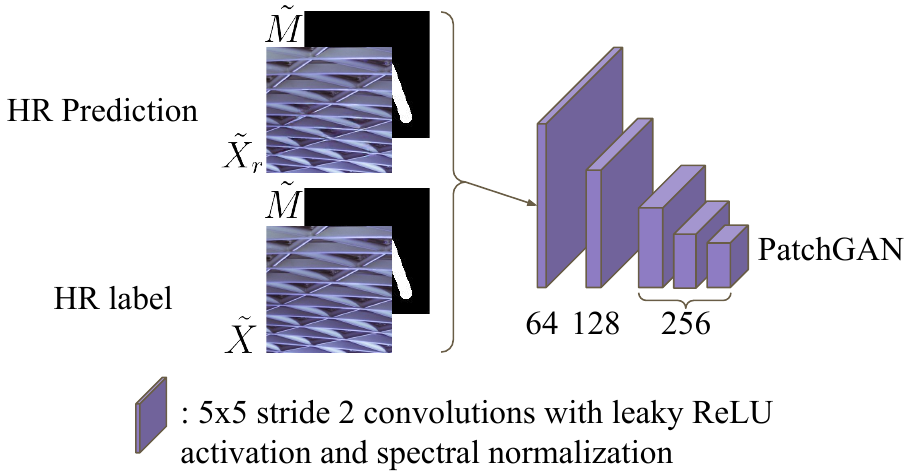}
\caption{Architecture of the PatchGAN discriminator used during training.}
\label{fig:d}
\end{figure}

\begin{figure*}
\centering
\includegraphics[width=\textwidth]{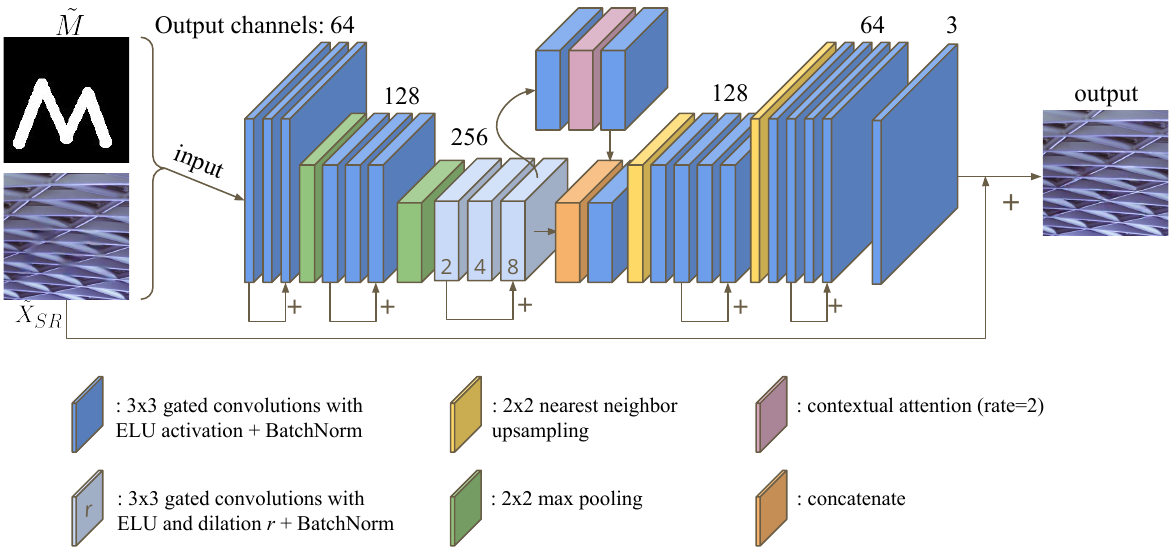}
\caption{Architecture of the refinement network in our zoom-to-inpaint framework.}
\label{fig:refine_net}
\end{figure*}

\subsection{Refinement network}
The architecture of our refinement network is shown in \fig{fig:refine_net}. The refinement network is similar to the coarse network except that a contextual attention \cite{yu2018generative} module is added to the bottleneck. The contextual attention module copies patches from the surrounding known regions into the regions to be inpainted based on the computed similarity. For memory-wise efficiency, the attention map is computed on $\times1/2$ downscaled feature maps of the bottleneck, and applied on the original resolution of the bottleneck features. The resulting feature maps after contextual attention are concatenated with the feature maps of the main pass at the bottleneck so that proceeding layers can merge both information sources as needed. In our complete zoom-to-inpaint framework, the refinement network works at high resolution (HR) of $512\times 512$ for enhanced refinement with magnification (zoom). Note that the output of the refinement network is blended with the HR label when calculating the losses as mentioned in \eq{eq:refinement_output}.

\subsection{Discriminator}
The network architecture of the discriminator used for training is shown in \fig{fig:d}. We use a hinge GAN loss with a PatchGAN \cite{isola2017imagetoimage} discriminator that aims to discriminate between the HR label and the HR prediction generated by the refinement network (blended with the HR label). $5\times 5$ convolution filters with stride 2, and leaky ReLU \cite{maas2013rectifier} activation are used. Spectral normalization \cite{miyato2018spectral} is applied at every layer for stable training of the GAN framework.

\begin{table*}
    \centering
    \scalebox{0.77}{
    \begin{tabular}{c|c|c|c|c|c|c}
    \hline\hline
        \multirow{2}{*}{Mask} & Number of & Length of & Thickness & \multirow{2}{*}{Angle $a$ for each vertex} & Every other angle & Invalid pixel ratio (mean / max \\
        & vertices & piece-wise stroke & of stroke & & for each vertex & from 10K random masks) \\
        \hline\hline
        $n=1$ & $\mathcal{U}\{1, 12\}$ & $\mathcal{U}\{1, d/12\}$ & $\mathcal{U}\{5, 30\}$ & $\mathcal{U}\{0, 2\pi\}$ & $a+\mathcal{U}\{7\pi/8, 9\pi/8\}$ & 5.61\% / 19.16\%\\
        \hline
        $n=2$ & $\mathcal{U}\{4, 12\}$ & $\mathcal{N}\{d/8, d/16\}$ & $\mathcal{U}\{12, 40\}$ & $2\pi/5+\mathcal{U}\{\mathcal{U}\{-2\pi/15, 0\}, \mathcal{U}\{0, 2\pi/15\}\}$ & $2\pi-a$ & 15.12\% / 50.24\%\\
        \hline\hline
    \end{tabular}}
    \caption{Probability distributions of mask generation parameters for masks at $n=1, 2$. $d$ denotes the diagonal length of the image.
    }
    \label{tab:mask}
\end{table*}

\begin{table*}
    \centering
    \scalebox{0.9}{
    \begin{tabular}{c|c|c|c|c|c}
    \hline\hline
    Method & HiFill \cite{yi2020contextual} & Pluralistic \cite{zheng2019pluralistic} & DeepFill-v2 \cite{yu2019free} & EdgeConnect \cite{nazeri2019edgeconnect} & Ours\\
    \hline\hline
    Number of Parameters & 2.7M* & 6M* & 4.1M* & 21.53M & 4.5M\\
    \hline
    Inference Time & 127 ms (60 ms) & 37 ms & 71 ms & 21 ms & 293 ms\\
    \hline\hline
    \multicolumn{6}{l}{*Values copied from the publication.}
    \end{tabular}}
    \caption{Comparison of the number of parameters and average inference time in milliseconds (ms) on 100 images. The inference time for HiFill is measured on crops enlarged to $512\times 512$. Other methods are measured for $256\times 256$ crops. Value in brackets denotes the time measured without pre- and post-processing for HiFill.}
    \label{tab:inference}
\end{table*}

\section{Mask generation}
\tab{tab:mask} shows a detailed configuration of the parameters used for generating masks at $n=1$ (\textit{masks}) and $n=2$ (\textit{large masks}). Each mask generation parameter is drawn from a probability distribution shown in \tab{tab:mask}, and a random mask is generated using those parameters following the mask generation scheme in DeepFill-v2 \cite{yu2019free}. A mask is generated by drawing random strokes on a $256\times 256$ tensor filled with zeros. First, a random starting location $(x, y)$ and a random number of vertices is chosen for each stroke. Then for each stroke, a piece-wise stroke is drawn between the vertices with a randomly chosen length of piece-wise stroke (distance between two vertices) and a random angle. A random thickness is fixed for each stroke. The \textit{large masks} ($n=2$) are generated with the original parameters used in \cite{yu2019free}, and \textit{masks} ($n=1$) are generated using adjusted parameters, modified so that the invalid pixel ratio is smaller and the generated masks are more confined.

\section{Complexity analysis}
We compare the number of parameters and inference time of HiFill \cite{yi2020contextual}, Pluralistic \cite{zheng2019pluralistic}, DeepFill-v2 \cite{yu2019free}, EdgeConnect \cite{nazeri2019edgeconnect} and Ours in \tab{tab:inference}. The average inference time in milliseconds (ms) is measured over 100 center crops of size $256\times 256$ of the DIV2K validation dataset for Pluralistic \cite{zheng2019pluralistic}, DeepFill-v2 \cite{yu2019free}, EdgeConnect \cite{nazeri2019edgeconnect} and Ours, which were all trained on $256\times 256$ images. For HiFill \cite{yi2020contextual}, which was trained on $512\times 512$, we enlarge the mask by nearest neighbor upsampling and the image by bicubic upsampling to match the resolution it was trained on, and measure the inference time, same as the way we performed the qualitative and quantitative evaluations. 

Image file I/O (read and write) times are excluded and we only measure the time it takes to run inference on the CNN for all methods except HiFill. For HiFill, we measured the inference time both with and without pre- and post-processing, which includes their proposed residual aggregation module. For our model, we measure the time it takes to process the input image through the entire pipeline including the coarse network, SR network, refinement network and bicubic downscaling. Because our refinement network handles higher-resolution images to generate better high-frequency details, the inference time is longer compared to other inpainting methods. However, an average inference time of 293 ms still allows interactive inpainting with users. The inference time was measured on an NVIDIA Tesla T4 GPU.

\begin{table*}
    \centering
    \scalebox{0.84}{
    \begin{tabular}{c|c|c|c|c|c|c|c|c|c|c}
    \hline\hline
         \multirow{2}{*}{Ablations} & \multirow{2}{*}{ch.} & \multicolumn{4}{c|}{\textit{Masks}} & \multicolumn{4}{c|}{\textit{Large Masks}} & Number of \\
         \cline{3-10}
          & & PSNR $\uparrow$ & SSIM $\uparrow$ & MS-SSIM $\uparrow$ & L1 Error $\downarrow$ & PSNR $\uparrow$ & SSIM $\uparrow$ & MS-SSIM $\uparrow$ & L1 Error $\downarrow$ & Parameters \\
         \hline\hline
         No zoom & 64 & 32.12 & 0.9714 & 0.9812 & 0.00441 & 25.89 & 0.8977 & 0.9180 & 0.01747 & 4.03M \\
         \hline
         Bicubic zoom & 64 & 32.80 & 0.9753 & 0.9832 & 0.00391 & 26.09 & 0.9016 & 0.9219 & 0.01657 & 4.03M \\
         \hline\hline
         No zoom & 70 & 32.72 & 0.9723 & 0.9818 & 0.00432 & 26.00 & 0.8993 &  0.9195 & 0.01761 & 4.82M \\
         \hline
         Bicubic zoom & 70 & 32.78 & 0.9730 & 0.9833 & 0.00413 & 25.99 & 0.9000 & 0.9212 & 0.01685 & 4.82M \\
         \hline\hline
         SR zoom & 64 & 33.40 & 0.9770 & 0.9853 & 0.00363 & 26.95 & 0.9080 & 0.9307 & 0.01497 & 4.48M \\
         \hline
         SR zoom+$L_\nabla$ & 64 & \textbf{34.08} & \textbf{0.9787} & \textbf{0.9886} & \textbf{0.00329} & \textbf{27.07} & \textbf{0.9094} & \textbf{0.9346} & \textbf{0.01462} & 4.48M \\
         \hline\hline
    \end{tabular}}
    \caption{Extended ablation study results including models with 70 output channels for \textit{No zoom} and \textit{Bicubic zoom}. The \textit{SR zoom} models still outperform \textit{No zoom} and \textit{Bicubic zoom} models even with less number of trainable parameters, signifying that the benefits of \textit{SR zoom} come from the framework design rather than network capacity. Values in \textbf{bold} denote best performance and ch. denotes the number of output channels at each convolution layer.}
    \label{tab:ablation_supp}
\end{table*}

\begin{table*}
    \centering
    \scalebox{0.85}{
    \begin{tabular}{c|c|c|c|c|c|c|c}
    \hline\hline
    \multirow{2}{*}{Frequency Range} & \multicolumn{4}{c|}{Models} & \multicolumn{3}{c}{Difference} \\
    \cline{2-8}
    & (a) No zoom & (b) Bicubic zoom & (c) SR zoom & (d) SR zoom+$L_\nabla$ & (b)-(a) & (c)-(a) & (d)-(a)\\
    \hline\hline
    Low Frequency & \textbf{0.9858} & \textbf{0.9870} & \textbf{0.9888} & \textbf{0.9907} & 0.0012 & 0.0030 & 0.0049 \\
    \hline
    Mid Frequency & 0.9730 & 0.9756 & 0.9772 & 0.9793 & 0.0026 & 0.0042 & 0.0063 \\
    \hline
    High Frequency & 0.9674 & 0.9717 & 0.9727 & 0.9743 & \textbf{0.0043} & \textbf{0.0053} & \textbf{0.0069} \\
    \hline\hline
    \end{tabular}}
    \caption{SSIM values measured for different frequency ranges using Laplacian pyramids in \cite{burt1983laplacian} on ablation models. It can be observed that the SSIM gain increases for higher-frequency components on models (b), (c) and (d) over (a). The difference in SSIM is plotted as a bar graph in \fig{fig:freq_analysis}. Highest values denoted in \textbf{bold}.}
    \label{tab:frequency}
\end{table*}

\section{Extended quantitative evaluations}
\subsection{Ablation study}
We provide an extended ablation study result in \tab{tab:ablation_supp} on $256\times256$ center crops of the DIV2K validation set on both mask sizes. It includes the four ablation models introduced in the main paper: (i) No zoom, (ii) Bicubic zoom, (iii) SR zoom, and (iv) SR zoom+$L_\nabla$. For (i) No zoom, we replace the SR component with an identity transform that simply copies the coarse output without an upsampling component, so that refinement is applied at the original resolution like in other
conventional 2-stage inpainting frameworks. For (ii) Bicubic zoom, we replace the SR component with bicubic upsampling, and for (iii), we add back the SR zoom. (i), (ii) and (iii) are trained without the gradient loss, $L_\nabla$. Lastly, (iv) SR zoom+$L_\nabla$ is our final framework with gradient loss. 

The SR zoom models, (iii) and (iv), contain an additional trainable component -- the SR network. These models have 4.48M trainable parameters, compared to No zoom and Bicubic zoom with 4.03M trainable parameters at the same number of 64 output channels. To test the effect of the network capacity on performance, we increase the number of output channels from 64 to 70 in the coarse network and the refinement network of the No zoom and Bicubic zoom frameworks so that they have more parameters (4.82M) than the SR zoom models. The quantitative results along with the number of parameters are provided in \tab{tab:ablation_supp}. As shown in \tab{tab:ablation_supp}, the \textit{SR zoom} models still outperform the \textit{No zoom} and \textit{Bicubic zoom} models even with less number of parameters and we can conclude that the benefits of our SR zoom framework are not from the increased number of parameters.

\begin{table}
    \centering
    \scalebox{0.8}{
    \begin{tabular}{c|c|c|c|c}
         \multirow{2}{*}{Method} & \multicolumn{4}{c}{Places2 ($256\times256$)} \\
         \cline{2-5}
          & PSNR $\uparrow$ & SSIM $\uparrow$ & MS-SSIM $\uparrow$ & L1 Error $\downarrow$\\
         \hline
         \rowcolor[gray]{.95}\multicolumn{5}{c}{Values copied from the publications} \\
         \hline
         HiFill \cite{yi2020contextual} & - & - & \textcolor{mygray}{0.8840} & \textcolor{mygray}{0.05439} \\
         \hline
         Pluralistic \cite{zheng2019pluralistic} & - & - & - & - \\
         \hline
         DeepFill-v2 \cite{yu2019free} & - & - & - & \textcolor{mygray}{0.09100} \\
         \hline
         EdgeConnect \cite{nazeri2019edgeconnect} & \textcolor{mygray}{27.95} & \textcolor{mygray}{0.9200} & - & \textcolor{mygray}{0.01500} \\
         \multicolumn{5}{c}{\,} \\
         \hline
         \rowcolor[gray]{.95}\multicolumn{5}{c}{Our measurements -- \textit{Masks}} \\
         \hline
         HiFill \cite{yi2020contextual} & 31.12 & 0.9586 & 0.9742 & 0.00744 \\
         \hline
         Pluralistic \cite{zheng2019pluralistic} & 33.23 & 0.9670 & 0.9807 & 0.00558 \\
         \hline
         DeepFill-v2 \cite{yu2019free} & 34.03 & 0.9719 & 0.9834 & 0.00485 \\
         \hline
         EdgeConnect \cite{nazeri2019edgeconnect} & 33.98 & 0.9718 & 0.9841 & 0.00388 \\
         \hline\hline
         Ours & \textbf{34.78} & \textbf{0.9755} & \textbf{0.9863} & \textbf{0.00357} \\
         \multicolumn{5}{c}{\,} \\
         \hline
         \rowcolor[gray]{.95}\multicolumn{5}{c}{Our measurements -- \textit{Large Masks}} \\
         \hline
         HiFill \cite{yi2020contextual} & 24.94 & 0.8891 & 0.9134 & 0.02034 \\
         \hline
         Pluralistic \cite{zheng2019pluralistic} & 26.17 & 0.9022 & 0.9191 & 0.01784 \\
         \hline
         DeepFill-v2 \cite{yu2019free} & 26.77 & 0.9158 & 0.9326 & 0.01536 \\
         \hline
         EdgeConnect \cite{nazeri2019edgeconnect} & 27.61 & 0.9166 & 0.9382 & 0.01328 \\
         \hline\hline
         Ours & \textbf{27.71} & \textbf{0.9202} & \textbf{0.9415} & \textbf{0.01314}\\
    \end{tabular}}
    \caption{Extended quantitative comparison on Places2.  Values copied from the original publications are denoted in gray for reference, but they are not directly comparable to our measurements as they were all tested under different settings, as summarized in \tab{tab:quantitative2}. Values in \textbf{bold} denote best performance among our measurements.}
    \label{tab:quantitative_supp}
\end{table}

\subsection{Numerical values for frequency analysis}
In \tab{tab:frequency}, we provide the raw numerical values of SSIM used for plotting the bar graph in \fig{fig:freq_analysis}. For this experiment, Laplacian pyramids \cite{burt1983laplacian} were constructed for the four ablation models ((a) No zoom, (b) Bicubic zoom, (c) SR zoom, (d) SR zoom+$L_\nabla$) and the ground truth using a traditional 5-tap Gaussian kernel, and SSIM values of the ablation models were measured for each frequency range to examine the performance gain at different frequency ranges. Low, mid and high frequency ranges each correspond to level 2, 1 and 0 in the Laplacian pyramid, respectively, with the last level (level 2) being the remaining blurred image.  As shown in \tab{tab:frequency}, the absolute SSIM values tend to be higher for lower frequencies, which are easier to reconstruct. However, the relative SSIM gain of (b), (c) and (d) over (a) is higher for higher frequencies, showing the benefits of the HR refinement models in generating high-frequency components in the inpainted results.

\begin{table*}
    \centering
    \scalebox{0.9}{
    \begin{tabular}{c|c}
    \hline\hline
    Method & Testing conditions in original publications \\
    \hline\hline
    \multirow{2}{*}{HiFill \cite{yi2020contextual}} & Tested on Places2 validation set, randomly cropped by $512\times 512$.\\
    & Used random masks from \cite{liu2018image} as well as their own random object masks.\\
    \hline
    \multirow{2}{*}{Pluralistic \cite{zheng2019pluralistic}} & Only provided quantitative evaluations on ImageNet ($256\times256$). \\
    & Proposed own random masks with random lines, circles and ellipses.\\
    \hline
    \multirow{2}{*}{DeepFill-v2 \cite{yu2019free}} & Tested on Places2 validation set ($256\times256$). Proposed random brush stroke masks \\
    &that can be generated on-the-fly during training (same as our \textit{large masks}).\\
    \hline
    \multirow{3}{*}{EdgeConnect \cite{nazeri2019edgeconnect}} & \multirow{3}{*}{\shortstack{Tested on Places2 ($256\times 256$) but no mention on which images were used.\\Used random masks from \cite{liu2018image}.}}\\
     & \\
    & \\
    \hline\hline
    \end{tabular}}
    \caption{Descriptions on quantitative evaluations performed in the original publications, which were used for obtaining the values in gray in \tab{tab:quantitative_supp} that were copied from the original publications.}
    \vspace{-1mm}
    \label{tab:quantitative2}
\end{table*}

\begin{table}
    \centering
    \scalebox{0.83}{
    \begin{tabular}{c|c|c|c|c}
         \multirow{2}{*}{Method} & \multicolumn{2}{c|}{Places2 ($256\times256$)} & \multicolumn{2}{c}{DIV2K ($256\times256$)}\\
         \cline{2-5}
          & LPIPS $\downarrow$ & FID $\downarrow$ & LPIPS $\downarrow$ & FID $\downarrow$\\
         \hline
         \rowcolor[gray]{.95}\multicolumn{5}{c}{\textit{Masks}} \\
         \hline
         HiFill \cite{yi2020contextual} & 0.0306 & 24.71 & 0.0258 & 12.85 \\
         \hline
         Pluralistic \cite{zheng2019pluralistic} & 0.0236 & 82.43 & 0.0219 & 128.21 \\
         \hline
         DeepFill-v2 \cite{yu2019free} & 0.0178 & 16.58 & 0.0172 & 8.61 \\
         \hline
         EdgeConnect \cite{nazeri2019edgeconnect} & 0.0177 & 16.41 & 0.0171 & 8.51 \\
         \hline\hline
         Ours & \textbf{0.0175} & \textbf{14.32} & \textbf{0.0147} & \textbf{7.14} \\
         \multicolumn{5}{c}{\,} \\
         \hline
         \rowcolor[gray]{.95}\multicolumn{5}{c}{\textit{Large Masks}} \\
         \hline
         HiFill \cite{yi2020contextual} & 0.0889 & 56.86 & 0.0989 & 56.65 \\
         \hline
         Pluralistic \cite{zheng2019pluralistic} & 0.0745 & 62.73 & 0.0810 & 150.28 \\
         \hline
         DeepFill-v2 \cite{yu2019free} & 0.0585 & 38.22 & 0.0636 & 33.89 \\
         \hline
         EdgeConnect \cite{nazeri2019edgeconnect} & \textbf{0.0554} & \textbf{35.41} & \textbf{0.0611} & \textbf{32.57} \\
         \hline\hline
         Ours & 0.0603 & 38.85 & 0.0628 & 42.25 \\
    \end{tabular}}
    \caption{Quantitative comparison using perceptual metrics, LPIPS and FID. Values in \textbf{bold} denote best performance.}
    \vspace{-2mm}
    \label{tab:perceptual}
\end{table}

\subsection{Quantitative comparison}
We provide an extended table on the quantitative comparison with other inpainting methods \cite{nazeri2019edgeconnect, yi2020contextual, yu2019free, zheng2019pluralistic} in \tab{tab:quantitative_supp}. For reference, in addition to the values measured on our Places2 test set, which were also reported in the main paper, we have added the metric values from the original publications measured on Places2. Note that they were evaluated under different settings in their original papers, as summarized in \tab{tab:quantitative2}, and are not directly comparable to the values we have measured on \textit{masks} and \textit{large masks}. Furthermore, we provide a comparison on perceptual metrics -- FID \cite{fid} and LPIPS \cite{lpips} in \tab{tab:perceptual}. Our model outperforms all other methods for \textit{masks}, and is 2nd or 3rd best for \textit{large masks}. The results on perceptual metrics seem to correspond to those of the user study shown in \sect{sect:user_study}, where we also provide an analysis on \textit{large masks}.

\begin{figure}
\centering
\includegraphics[width=\columnwidth]{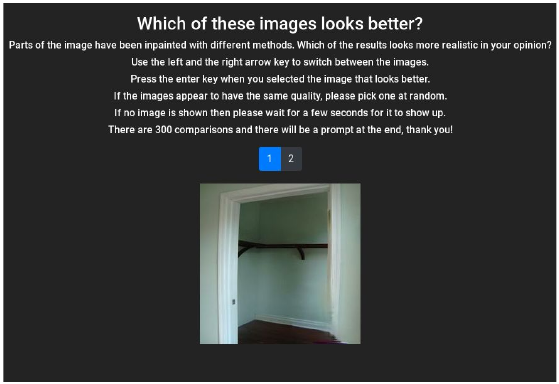}
\caption{Screenshot of user study presented to participants.}
\vspace{-1mm}
\label{fig:user_study}
\end{figure}

\section{User study}
\label{sect:user_study}
As mentioned in the main paper, we conducted a user study, in which 13 users evaluated the inpainting results. Each user compared 300 image pairs, where one is generated by our method and the other is generated by one of the inpainting methods in \cite{nazeri2019edgeconnect, yi2020contextual, yu2019free, zheng2019pluralistic}. The appearing order of methods (whether ours comes first, or the compared method comes first) and the appearing order of image pairs was randomized for each user. A screenshot of the user study is shown in \fig{fig:user_study}. Participants could toggle between the two compared images and select a preferred version before proceeding to the next image pair for comparison. No time limitations were given. The raw number of counts obtained from the user study is shown in \tab{tab:count}, from which we calculated the percentage of users that preferred our method in \tab{tab:user_study}. 

\begin{table*}
    \centering
    \scalebox{0.9}{
    \begin{tabular}{c|c|c|c|c|c}
    & HiFill \cite{yi2020contextual} vs. Ours & Pluralistic \cite{zheng2019pluralistic} vs. Ours & DeepFill-v2 \cite{yu2019free} vs. Ours & EdgeConnect \cite{nazeri2019edgeconnect} vs. Ours & Total \\
    \hline
     \rowcolor[gray]{.95}\multicolumn{6}{c}{\textit{Masks}} \\
     \hline
    Counts & 239 vs. 736 & 106 vs. 869 & 300 vs. 675 & 349 vs. 626 & 3900 \\
    \hline
    Prefer Ours & 75.49\% & 89.13\% & 69.23\% & 64.21\% & 74.51\% \\
    \multicolumn{6}{c}{\,} \\
     \hline
     \rowcolor[gray]{.95}\multicolumn{6}{c}{\textit{Large Masks}} \\
     \hline
     Counts & 412 vs. 713 & 322 vs. 803 & 780 vs. 345 & 738 vs. 387 & 4500 \\
     \hline
     Prefer Ours & 63.38\% & 71.38\% & 30.67\% & 34.4\% & 49.96\% \\
    \end{tabular}}
    \caption{Raw counts of the user study results on \textit{masks} (top) and \textit{large masks} (bottom). The values at the top were used to compute the preference rate in \tab{tab:user_study}. 13 and 15 users participated in the user study with \textit{masks} and \textit{large masks}, respectively, and each user compared 300 image pairs.}
    \label{tab:count}
\end{table*}

\begin{figure*}
\centering
\includegraphics[width=\textwidth]{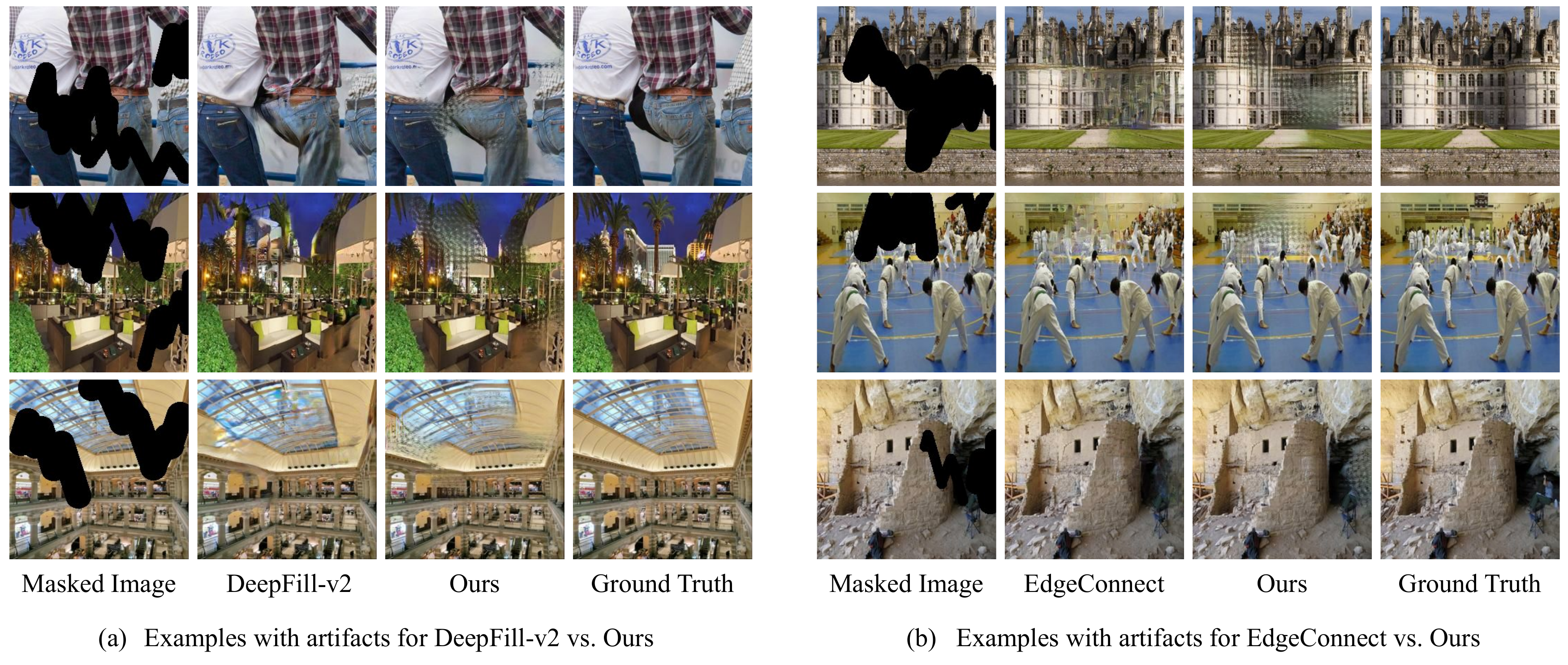}
\caption{Failure case examples with \textit{large masks} shown during the user study. As both the compared method and our method contain artifacts in these extreme cases with very large masks, it is difficult to compare the quality of the generated results, and users select images containing less objectionable artifacts. Our method tends to produce a repetitive high-frequency pattern that is displeasing, resulting in more users preferring DeepFill-v2 \cite{yu2019free} and EdgeConnect \cite{nazeri2019edgeconnect} for \textit{large masks}. Note that these are not average results and more typical results are shown in \fig{fig:qualitative_1} (b) and \fig{fig:qualitative} (b).}
\label{fig:artifacts}
\end{figure*}

\medskip\noindent
\textbf{Results on \textit{large masks} and failure case analysis.}\quad
In \tab{tab:count}, we show the results on \textit{large masks} that were additionally evaluated on another 15 users. Although our approach outperforms all other methods for \textit{masks}, for \textit{large masks}, more users preferred DeepFill-v2 \cite{yu2019free} and EdgeConnect \cite{nazeri2019edgeconnect}. As briefly explained in the Conclusion section in the main paper, some examples with \textit{large masks} contain sizeable missing regions, on which both the compared method and ours tend to generate severe artifacts. \fig{fig:artifacts} shows some of the example pairs (DeepFill-v2 vs Ours, or EdgeConnect vs Ours) containing artifacts that were presented to the users. In these extreme cases, it becomes difficult to compare the quality of the inpainted results, and users select the result with a less objectionable artifact. As shown in \fig{fig:artifacts}, our method tends to produce a repetitive high-frequency pattern that is displeasing in these cases. We consider this artifact pattern to be a highly likely reason as to why users preferred DeepFill-v2 or EdgeConnect over ours when the inpainted regions are very large. Note that these are failure cases on extreme cases, and not average results of our method on \textit{large masks}. More typical results on \textit{large masks} are shown in \fig{fig:qualitative_1} (b) and \fig{fig:qualitative} (b).

\begin{table*}
    \centering
    \scalebox{0.9}{
    \begin{tabular}{c|c|c|c|c|c|c|c|c|c}
    \hline\hline
         \multirow{2}{*}{Models} & \multirow{2}{*}{scale} & \multicolumn{4}{c|}{\textit{Masks}} & \multicolumn{4}{c}{\textit{Large Masks}} \\
         \cline{3-10}
          & & PSNR $\uparrow$ & SSIM $\uparrow$ & MS-SSIM $\uparrow$ & L1 Error $\downarrow$ & PSNR $\uparrow$ & SSIM $\uparrow$ & MS-SSIM $\uparrow$ & L1 Error $\downarrow$ \\
         \hline\hline
         No zoom & $\times1$ & 29.91 & 0.9382 & 0.9608 & 0.00955 & 23.54 & 0.8186 & 0.8555 & 0.03066\\
         \hline
         \multirow{3}{*}{SR zoom} & $\times2$ & \textbf{30.85} & \textbf{0.9439} & \textbf{0.9658} & \textbf{0.00855} & 24.23 & 0.8273 & 0.8700 & 0.02821\\
         \cline{2-10}
         & $\times3$ & 30.70 & 0.9433 & 0.9648 & 0.00868 & \textbf{24.36} & \textbf{0.8290} & \textbf{0.8710} & \textbf{0.02739}\\
         \cline{2-10}
         & $\times4$ & 29.92 & 0.9404 & 0.9625 & 0.00932 & 23.91 & 0.8265 & 0.8664 & 0.02963\\
         \hline\hline
    \end{tabular}}
    \caption{Experiment on different SR scale factors trained with $128\times128$ patches. The \textit{No zoom} model can be considered as a $\times1$ scale version of the framework. The best performing model is $\times2$ scale model for \textit{masks} and $\times3$ scale model for \textit{large masks}. Best values denoted in \textbf{bold}.}
    \label{tab:scale}
\end{table*}

\begin{figure*}
\centering
\includegraphics[width=\textwidth]{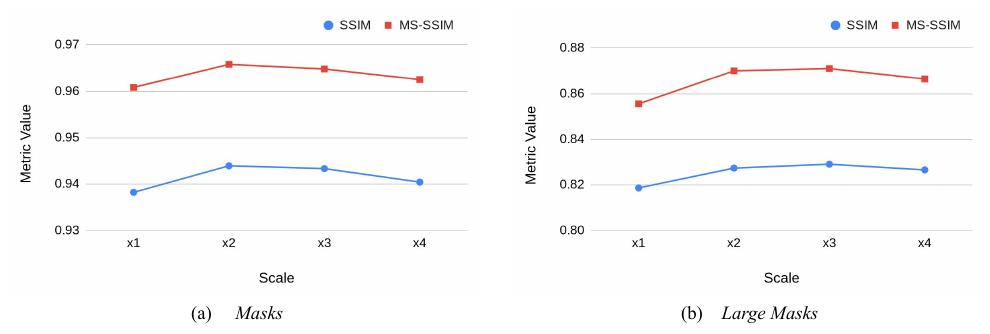}
\caption{Graphs showing SSIM and MS-SSIM performance plotted for models with scale factors $\times1$, $\times2$, $\times3$ and $\times4$, for (a) \textit{Masks} and (b) \textit{Large Masks}.}
\label{fig:scale}
\end{figure*}

\section{Analysis on different scale factors}
In the main paper, refinement is performed at double ($\times2$) the resolution of the coarse output. To analyze the effect of scale factors on our zoom-to-inpaint framework, we retrain the \textit{SR zoom} model (without gradient loss) on three different scale factors ($\times2$, $\times3$, $\times4$) and report the results on the validation set of DIV2K ($256\times256$ center crops) in \tab{tab:scale}. To avoid out-of-memory errors during training due to larger scale factors ($\times3$ and $\times4$ compared to $\times2$ in the original model), all models are trained on $128\times128$ patches instead of $256\times256$ in the original zoom-to-inpaint framework. As in the original framework, each component (coarse, SR and refinement network) is pretrained first, before the end-to-end progressive learning. We empirically find the loss coefficients ($\lambda$) on the model with $\times2$ and use the same values for models with $\times3$ and $\times4$. We also compare with the \textit{No zoom} model retrained on $128\times128$ patches, which can be considered as scale $\times1$. \fig{fig:scale} shows a graph with SSIM and MS-SSIM values plotted for the four models in \tab{tab:scale} for both mask sizes. As shown in \tab{tab:scale} and \fig{fig:scale}, all SR zoom models ($\times2$, $\times3$, $\times4$) outperform the No zoom model ($\times1$), demonstrating the benefits of refining at higher resolution. The best performing model is the $\times2$ scale model for \textit{masks}, and the $\times3$ scale model for \textit{large masks}. There is a trade-off with the scale factor increase, as more information can be learned from HR labels with higher scale factors possibly leading to better reconstruction quality, but at the same time, the refinement network has to handle larger holes, which becomes highly challenging. We expect there would also be a correlation with the network capacity when searching for an optimal scale factor for HR refinement, which would be related to the ability to learn from HR labels.

\section{Additional visual results}
\subsection{Comparison to other inpainting methods}
\begin{figure*}
\centering
\includegraphics[width=\textwidth]{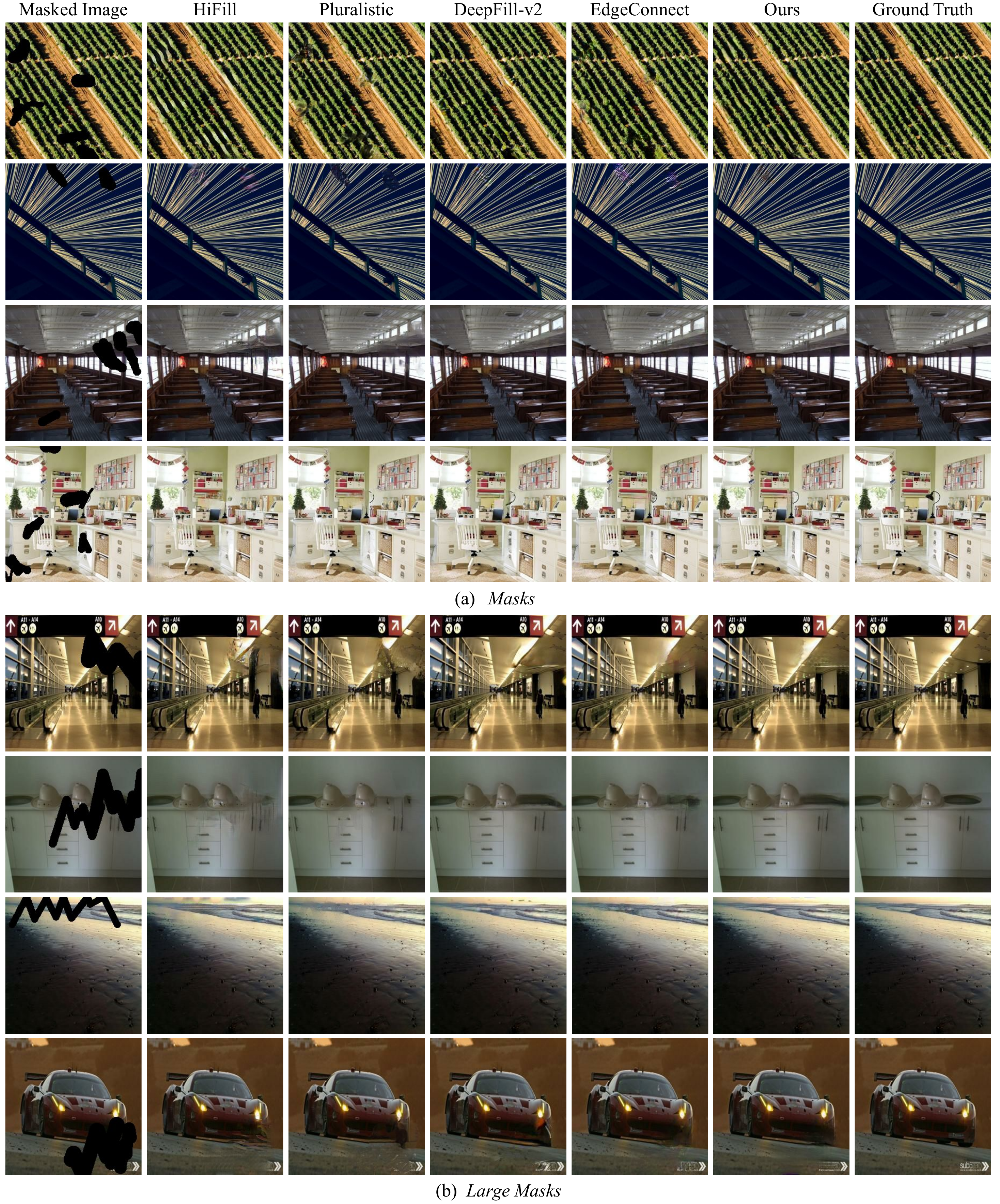}
\caption{Additional qualitative comparisons to HiFill \cite{yi2020contextual}, Pluralistic \cite{zheng2019pluralistic}, DeepFill-v2 \cite{yu2019free} and EdgeConnect \cite{nazeri2019edgeconnect} on (a) \textit{Masks} and (b) \textit{Large Masks}. Our zoom-to-inpaint model is able to reconstruct natural structures as shown in the examples in the 1st, 5th and 6th rows, and generate high-frequency components such as fine edges as shown in 2nd and 3rd rows.}
\label{fig:qualitative}
\end{figure*}

We provide additional comparisons to existing inpainting methods, HiFill \cite{yi2020contextual}, Pluralistic \cite{zheng2019pluralistic}, DeepFill-v2 \cite{yu2019free} and EdgeConnect \cite{nazeri2019edgeconnect}, in \fig{fig:qualitative}. Four rows at the top are results on \textit{masks}, and four rows at the bottom show results on \textit{large masks}. Our zoom-to-inpaint model is able to generate accurate structure information (1st row, 5th row, 6th row) as well as high-frequency details (2nd row, 3rd row).

\subsection{Ablation models}
\begin{figure*}
\centering
\includegraphics[width=\textwidth]{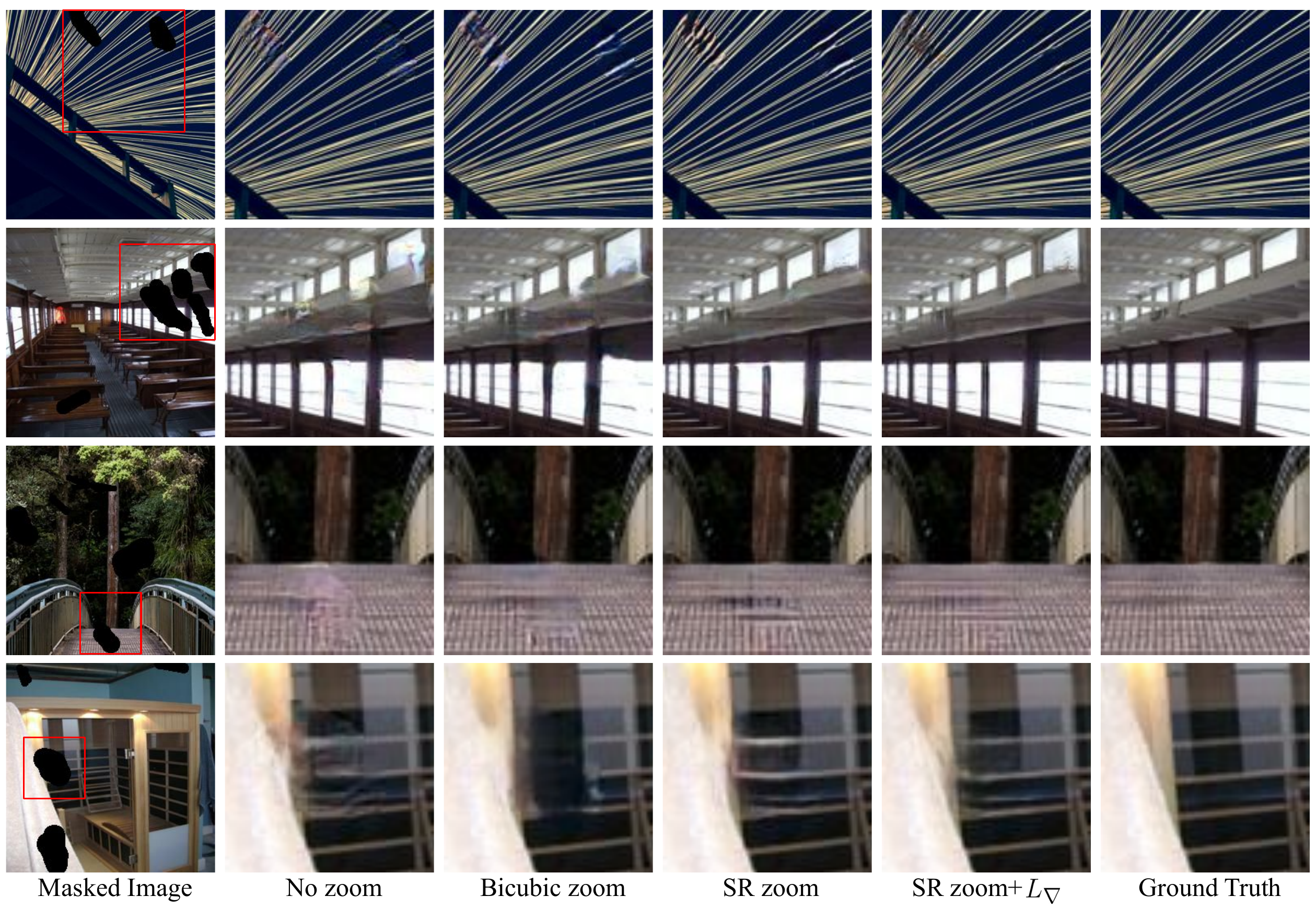}
\caption{Additional visual comparison on results generated by the ablation models. SR zoom helps improve the generation of high-frequency components, and gradient loss ($L_\nabla$) enhances the details even further.}
\label{fig:ablation_supp}
\end{figure*}

In \fig{fig:ablation_supp}, we further provide additional results generated by the four ablation models: No zoom, Bicubic zoom, SR zoom and SR zoom+$L_\nabla$. As shown, \textit{SR zoom} improves the generation of high-frequency details compared to \textit{No zoom} and \textit{Bicubic zoom} models, and gradient loss improves them even further. 

\subsection{Intermediate results}
\begin{figure*}
\centering
\includegraphics[width=\textwidth]{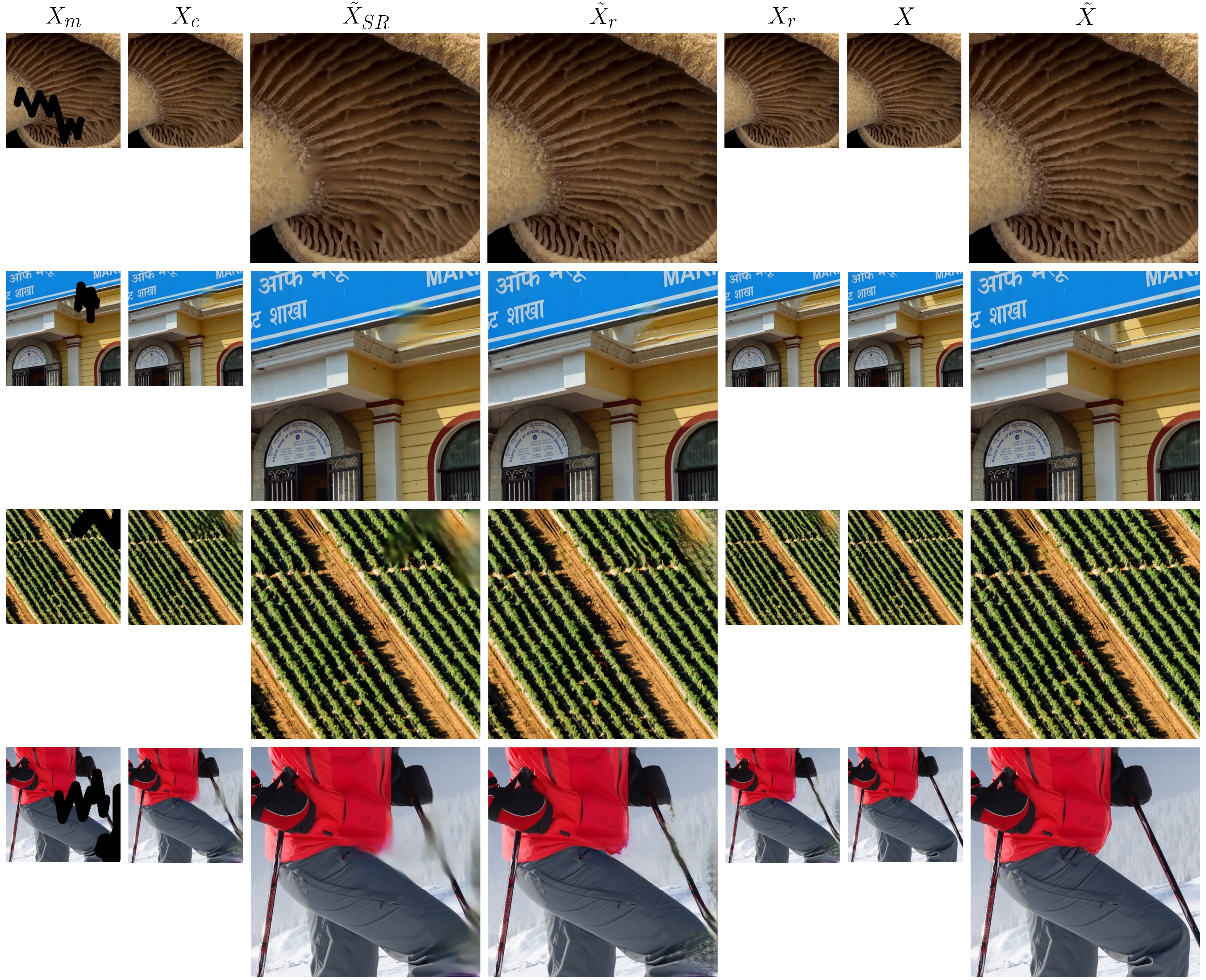}
\caption{Intermediate results extracted from our framework.}
\label{fig:intermediate}
\end{figure*}

Our zoom-to-inpaint model is a 4-stage framework, and the intermediate output of each stage can be extracted. \fig{fig:framework} shows the actual intermediate results produced by our framework. In \fig{fig:intermediate}, we provide additional examples of intermediate results, specifically $X_m$, $X_c$, $\tilde{X}_{SR}$, $\tilde{X}_r$ and $X_r$, along with LR and HR labels $X$ and $\tilde{X}$.

\end{appendices}
\end{document}